\documentclass{article}

\usepackage{PRIMEarxiv}

\usepackage[utf8]{inputenc} 
\usepackage[T1]{fontenc}    
\usepackage{hyperref}       
\usepackage{url}            
\usepackage{booktabs}       
\usepackage{amsfonts}       
\usepackage{nicefrac}       
\usepackage{microtype}      
\usepackage{lipsum}
\usepackage{fancyhdr}       
\usepackage{graphicx}       
\usepackage[table]{xcolor}
\graphicspath{{media/}}     

\usepackage[utf8]{inputenc}
\usepackage[T1]{fontenc}
\usepackage{comment}
\usepackage{graphicx}
\usepackage{subcaption} 
\usepackage{makecell}  
\usepackage{siunitx}

\usepackage{multirow}
\usepackage{listings}
\usepackage{verbatim}

\usepackage{fancyvrb}
\DefineVerbatimEnvironment{BoldVerbatim}{Verbatim}{
  formatcom=\bfseries\ttfamily,
  fontsize=\small              
}

\newcommand{\m}[1]{\textcolor{black}{#1}}
\newcommand{\ma}[1]{\textcolor{black}{#1}}

\title{Sensory-Motor Control with Large Language Models via Iterative Policy Refinement}

\author{
  Jônata Tyska Carvalho \\
  Federal University of Santa Catarina (UFSC) \\
  Florianópolis, Brazil\\
  Institute of Cognitive Sciences and Technologies (ISTC-CNR) \\
  Rome, Italy\\
  \texttt{jonata.tyska@ufsc.br} \\
   \And
  Stefano Nolfi \\
  Institute of Cognitive Sciences and Technologies (ISTC-CNR) \\
  Rome, Italy\\
  \texttt{stefano.nolfi@cnr.it} \\
}

\begin{document}
\maketitle

\begin{abstract}
We propose a method that enables large language models (LLMs) to control embodied agents \m{through the generation of control policies that directly map} continuous observation vectors to continuous action vectors. At the outset, the LLMs generate a control strategy based on a textual description of the agent, its environment, and the intended goal. This strategy is then iteratively refined through a learning process in which the LLMs are repeatedly prompted to improve the current strategy, using performance feedback and sensory-motor data collected during its evaluation.
The method is validated on classic control tasks from the Gymnasium library and the inverted pendulum task from the MuJoCo library. The approach proves effective with relatively compact models such as GPT-oss:120b and Qwen2.5:72b. In most cases, it successfully identifies optimal or near-optimal solutions by integrating symbolic knowledge derived through reasoning with sub-symbolic sensory-motor data gathered as the agent interacts with its environment.
\end{abstract}


\section{Introduction}

Large Language Models (LLMs) have emerged as powerful tools for developing embodied agents capable of exhibiting goal-directed behaviors through interaction with external environments. This capability stems from their ability to acquire extensive knowledge and complex reasoning skills via observational learning, wherein they process and internalize patterns from large-scale textual and multimodal data. The knowledge and skills obtained in this manner extend beyond purely linguistic tasks and can be effectively repurposed to control embodied agents. This versatility positions LLMs as a promising foundation for training and deploying autonomous systems across a wide range of domains, including robotics, virtual assistants, and interactive simulations.

The straightforward approach to leveraging LLMs for embodied control involves equipping agents with a repertoire of motor primitives—basic motion control modules—that enable the execution of elementary actions required to achieve basic objectives, such as navigating to a target or grasping an object. In this approach, the LLM is prompted to generate descriptions with the sequence of necessary basic actions that permit the achievement of the goal. These descriptions are then translated into concrete behaviors by activating the corresponding sequence of motor primitives. For example, Huang et al. (2022)\cite{huang2022inner} showed that GPT-3~\cite{brown2020language} and Codex~ \cite{chen2021evaluating} could produce plausible-albeit sometimes impractical-action plans when provided with a task description and a single example of a suitable action plan.

Simple frameworks of this kind-relying solely on language-based instructions without incorporating environmental perception or the ability to observe change during action executions-can be extended and enhanced through the use of multimodal LLMs operating dynamically. These models are capable of processing visual data, either encoded as specialized embedding vectors~\cite{driess2023palm} or represented as textual descriptions generated by vision-language models~\cite{zeng2022comprehensive}. Furthermore, the framework can be improved by enabling the model to monitor the environment during action execution and to adjust its action plan as needed dynamically~\cite{wang24planning}.

However, frameworks based on predefined motor primitives face intrinsic limitations. The behavior of embodied agents often exhibits quasi-modular organization~\cite{west2003developmental,nolfi2021behavioral}, meaning that overall behavior can be decomposed into sub-behaviors or components. However, these components lack sharp boundaries and must be finely tuned to the characteristics of preceding and subsequent actions. Consequently, designing or training primitive behaviors independently can hinder behavioral integration and compromise overall performance~~\cite{nolfi2021behavioral}. Such approaches are, therefore, only suitable for domains characterized by low dynamical complexity and limited interdependence among sequential actions.

In this article, we explore an alternative approach in which LLMs generate \m{control strategies that produce} continuous action vectors directly, encoding the temporal evolution of the agent's actuator states and eliminating the need for predefined motor primitives. Furthermore, we investigate how such models can improve their ability to produce goal-directed behaviors through a learning process realized through iterated prompting. 

The remainder of the paper is organized as follows: Section 2 reviews related work; Section 3 describes our proposed method; Section 4 presents our experimental results; and Section 5 provides concluding remarks. A comprehensive account of the experimental setup, the obtained results, and the software used for replication and further experimentation is included in the appendix.


\section{Related work}

\ma{The most direct and widely adopted framework for sensory–motor learning is Reinforcement Learning (RL). In RL, an agent acquires a control policy by interacting with its environment and optimizing expected cumulative reward through evaluative feedback \cite{sutton2018reinforcement}. This paradigm has become a standard approach for learning continuous control policies in both simulated and real-world domains, including robotics, manipulation, and locomotion ~\cite{kober2013robotics_survey,lillicrap2015ddpg,haarnoja2018sac,kalashnikov2018qtopt}. RL provides a natural baseline for studying sensory–motor skill acquisition, as it directly addresses the problem of mapping sensory observations to motor actions through experience-driven optimization.}
 
\ma{Reinforcement learning encompasses a broad family of methods that differ along several dimensions. These include online versus offline (batch) learning, where policies are either optimized through ongoing interaction or learned solely from previously collected datasets ~\cite{levine2020offline_rl_tutorial,fu2020d4rl,kumar2020cql}, as well as model-free and model-based approaches, which respectively learn policies directly from experience or leverage an explicit model of the environment’s dynamics. Despite their differences, all RL variants ultimately rely on interaction data generated by embodied agents operating in their target environments, either directly during training or indirectly through datasets collected by some behavior policy. While this interaction-centric nature ensures task- and morphology-specific learning, it can also result in high interaction data requirements, particularly for complex control problems.}

\ma{In this work, we investigate LLM-based systems can support alternative forms of sensory–motor learning. Unlike standard RL, LLMs possess extensive prior knowledge and reasoning capabilities acquired through large-scale pretraining, which can be exploited to initialize the agent’s control policy and to propose structured variations that are likely to yield performance improvements. This could potentially accelerate the learning process and reduce the number of required interactions with the environment.}

As stated in the introduction, this article focuses on methods that utilize LLMs \m{to generate control policies that directly govern} the state of the agent's actuators, without relying on predefined motor primitives. Within this category, different approaches have been proposed that vary in terms of training requirements, the use of demonstration data, and the integration of iterative learning mechanisms. An initial method of this kind was developed by Brohan et al. (2023)~\cite{brohan2022rt}, who fine-tuned vision-language models (VLMs) to generate specialized action tokens that encode the state of an agent's actuators over time. The authors demonstrated that this method enables a robotic arm equipped with a camera to respond appropriately to task requests such as "Pick up the bag about to fall off the table" when the agent observes its own arm and several bags positioned on a table, or "Put the strawberry into the correct bowl" when it observes its arm, a strawberry, a bowl of bananas, and a bowl of strawberries.

More specifically, Brohan et al. trained the PaLI-X~\cite{chen2023pali} and PaLM-E~\cite{driess2023palm} models to respond to prompts formatted as: “What action should the robot take to [task instruction]?” paired with task-specific visual inputs. The models were trained to produce vectors of special action tokens that encode the gripper's extension and the angular positions of the robotic arm’s six joints over time, ensuring the execution of the desired behavior. These joint positions were discretized into 256 bins, each represented by a distinct action token.

Training data were collected from 13 robots operating in a kitchen-office environment over a 17-month period~\cite{brohan2022rt}. Each robot demonstration trajectory was annotated with a natural language description of the corresponding task. Beyond replicating the demonstrated behaviors, the trained models exhibited strong generalization capabilities, successfully performing novel tasks not encountered during training. However, a key limitation of this approach lies in its dependence on large-scale, labor-intensive demonstration datasets.

In contrast, other researchers have explored prompting large language models (LLMs) to generate action vectors directly without using fine-tuning. Wang et al. (2024)~\cite{wang24planning} employed GPT-4~\cite{bubeck2023sparks} to control a simulated quadruped robot with 12 degrees of freedom (DOF), enabling it to walk on a flat surface. The model received continuous observation vectors as input and produced corresponding action vectors at a frequency of 200 Hz. This was achieved by prompting GPT-4 with (i) a description of the robot and task, (ii) an explanation of the observation and action vectors format, and (iii) examples of correct observation and action vector pairs. This prompt induced in-context learning, enabling the model to generate effective walking actions in response to new observations.  

Notably, LLMs prompted in this manner can synthesize control strategies by integrating qualitatively distinct forms of information: (i) symbolic information conveyed through textual descriptions of the robot, environment, and task objectives, (ii) symbolic information produced via reasoning, and (iii) sub-symbolic information contained in continuous sensory-motor data.

Furthermore, Mirchandani et al. (2023)~\cite{mirchandani2023large} showed that LLMs can generate appropriate actions through in-context learning using only sensory-motor data, without receiving any explicit information about the robot, the environment, or the task objective. In their experiments, an LLM successfully controlled a Panda robotic arm~\cite{haddadin2024franka} in a “Marker in a Cup” task, relying solely on demonstration data. As in the previous work mentioned above, in-context learning enabled the model to produce appropriate action vectors in response to new observation vectors. 

Moreover, Mirchandani et al. (2023)~\cite{mirchandani2023large} demonstrated that LLMs can leverage their in-context learning capabilities to improve their control strategies over multiple learning cycles.  This was achieved by repeatedly prompting the model across several iterations and updating the demonstration data in each cycle. The updated dataset was composed of the best behavioral samples selected from both human demonstrations and samples generated by the models in previous iterations. 

This iterative learning process can be further enhanced by incorporating reward signals into the prompt—scalar values that quantify how well each demonstrated or generated behavior achieves the intended task objective.

In this article, we propose an LLM-based methodology \m{for generating and iteratively refining control strategies that directly govern the state of an agent’s actuators.} Our approach enables models to learn autonomously without relying on human-provided demonstrations. Instead, the model learns from a combination of (i) linguistic descriptions of the robot, its environment, and the task objectives and (ii) sensory-motor data and rewards generated throughout the learning process.

Our approach integrates three key ideas: (i) leveraging LLMs' reasoning and in-context learning capabilities to derive control strategies from textual descriptions, as demonstrated by Wang et al. (2023)~\cite{wang24planning}; (ii) complementing this with learning from sensory-motor samples, as shown by Mirchandani et al. (2023)~\cite{mirchandani2023large}; and (iii) enabling iterative refinement of control strategies through in-context learning, as also proposed by Mirchandani et al. (2023)~\cite{mirchandani2023large},  However, in contrast to previous approaches, our method does not depend on a human demonstrator. Instead, it builds its knowledge entirely through autonomous experience, continuously improving its performance using data generated in previous iterations.

\ma{Recent work in LLM-based agentic control has also explored hybrid architectures that combine language-based reasoning with structured tools such as simulators, digital twins, or robotics~\cite{Kannan,Gill,vyas2025autonomous,ao2025llm}. In these approaches, LLMs typically function as planners or high-level policy generators, proposing sequences of tool invocations or the composition of predefined action primitives. While such methods are well suited for high-level reasoning, task decomposition, and domains where atomic actions or motor primitives are readily available, they are not designed to support low-level sensory–motor control.}

\ma{In contrast, our approach focuses on generating entire control programs offline, enabling efficient high-frequency execution without further LLM involvement during rollout. This design choice trades off dynamic adaptability for runtime efficiency but is mitigated through an iterative refinement loop that allows the model to update its policy based on execution feedback across episodes. Compared to purely zero-shot methods or direct observation-to-action mapping, our methodology offers a scalable and computationally efficient framework for control policy synthesis, particularly in settings where low latency and structured adaptation over time are important.}

The results obtained from classic control tasks in the Gymnasium library and the inverted pendulum problem in the MuJoCo library demonstrate that the proposed method consistently produces optimal or high-performing solutions. The analysis of the results underscores the value of combining the reasoning capabilities of large language models (LLMs) with a trial-and-error learning process driven by sensory-motor data and reward signals derived from the evaluation of candidate solutions. The approach proves effective with relatively compact models such as GPT-oss:120b and Qwen2.5:72b.

\section*{3. Method}

To evaluate the capability of LLMs to directly control the actuators of embodied agents, we developed a structured prompting technique. We assessed its effectiveness using classic control tasks from the Gymnasium library (formerly Gym ~\cite{gym:24}). The prompting methodology is described in Section 3.1, the control tasks considered are detailed in Section 3.2, the experimental conditions are outlined in Section 3.3, and the evaluation metrics are presented in Section 3.4. The code for replicating the experiments is publicly available together with all the supplementary material at https://github.com/jtyska/llm-robotics-article/.

\subsection*{3.1 Prompting technique}

The proposed methodology comprises two primary phases. In the first phase, the model is prompted to generate a control policy capable of achieving a desired behavior under varying environmental conditions. This is based on a textual description that characterizes the agent's body, sensors, actuators, environment, and task objectives. The description includes information on the agent's morphology and explains the structure and semantics of the values encoded in the observation and action vectors—i.e., the sensory inputs and motor outputs. 

In the second phase, the model is repeatedly prompted to refine its policy using the description of the performance outcomes of current and previous policies and sensory-motor data generated during policy evaluation.  

To promote structured reasoning and ensure the synthesis of effective control policies, we adopted a step-by-step prompting strategy. First, the LLM is asked to reflect on the problem and propose a high-level control strategy (Prompt 1, Figure 1). Next, it is instructed to translate this strategy into a set of IF-THEN-ELSE rules (Prompt 2). Finally, these rules are converted into executable Python code (Prompt 3). Preliminary experiments in which the LLM was directly instructed to generate Python code-without first articulating a high-level strategy and formulating it in a set of unambiguous rules-resulted in poor-quality output (data not shown), underscoring the importance of the structured prompting methodology.

Following the generation of the initial policy, the model undergoes an iterative improvement phase. It is prompted to revise its policy based on the same initial information, \m{augmented with descriptions of the current, previous, and best control policies generated so far (when available), along with their associated performance scores and the sensory-motor data collected during the evaluation of the current policy }(Prompt 4). This iterative refinement allows the model to adjust its control strategy and observe the impact of each modification. Due to token length constraints inherent in LLMs, the sensory-motor data included in each prompt are limited to the final 20 steps of a single evaluation episode.

An overview of this prompting methodology is illustrated in Figure 1, while the full prompt texts are provided in Appendix A.

\begin{figure}[!htbp]
\centering
\includegraphics[width=\linewidth]{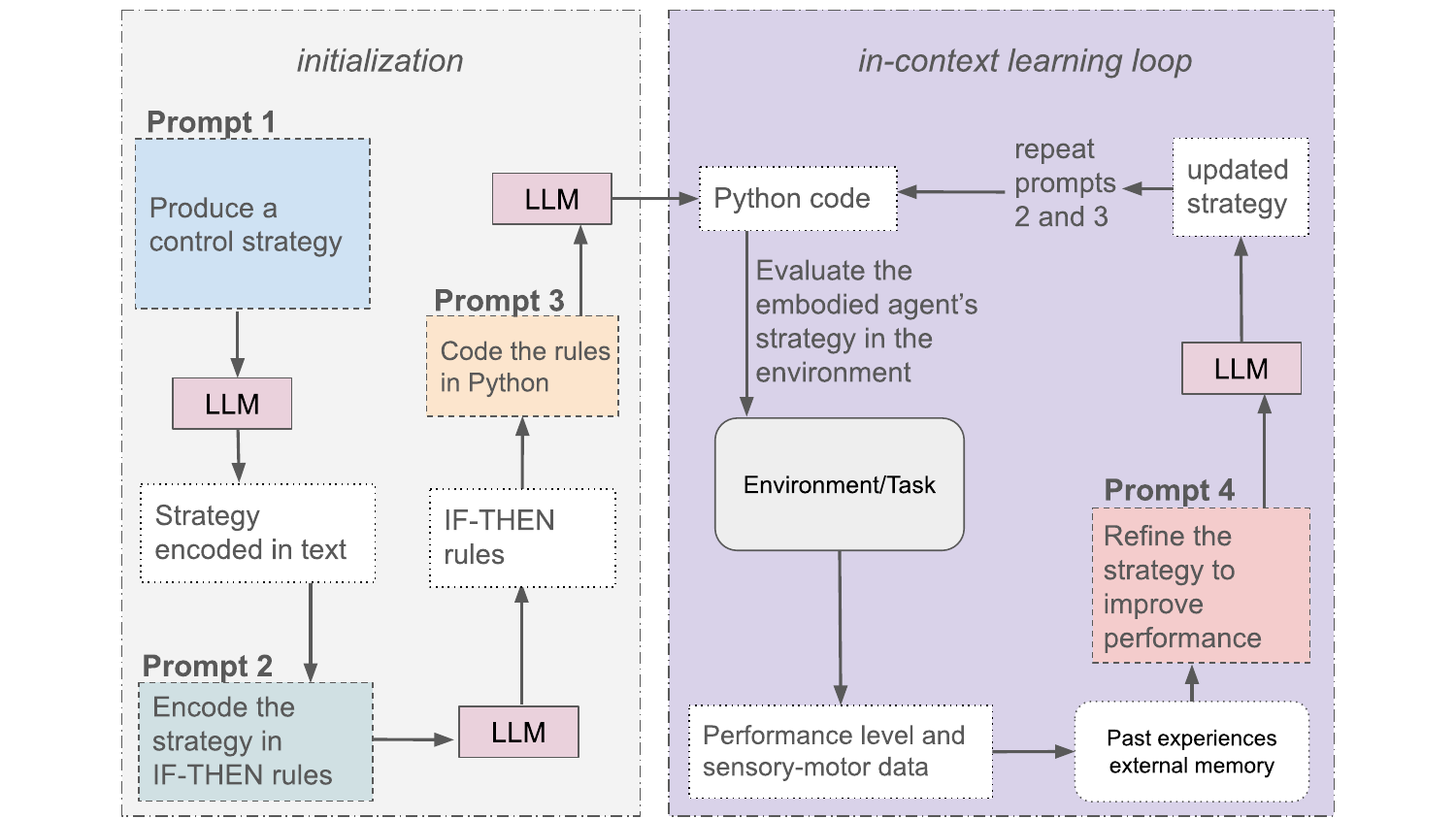}
\caption{Overview of the prompting method enabling LLMs to directly control the actuators of embodied agents.}
\label{fig:overview}
\end{figure}

It is important to emphasize that the agent and task descriptions provided in the prompts do not include all the details necessary to generate a complete solution. For instance, in one of the tasks we used, the pole balancing, critical parameters such as pole length and torque magnitude were deliberately omitted. This omission illustrates a fundamental challenge in embodied control compared to mathematical or abstract reasoning tasks: the number of relevant variables is substantially larger, making it generally impractical to specify every detail of the control problem explicitly.  

Nevertheless, our experiments demonstrate that such missing information can be inferred during the iterative learning process through the analysis of sensory-motor data and the observed outcomes of alternative control strategies. This capacity to integrate symbolic knowledge—conveyed through pre-trained data, textual prompts, and reasoning—with sub-symbolic knowledge—extracted from interaction with the environment—is central to the effectiveness of LLMs in embodied control tasks. It enables the derivation of effective solutions even in the presence of incomplete problem specifications.

Our proposed methodology facilitates problem-solving by drawing on three complementary sources of knowledge: (i) the general knowledge and reasoning abilities of LLMs, (ii) insights acquired through trial-and-error learning, and (iii) reasoning based on the observed effects of incremental policy modifications. This combination supports the generation of initial control strategies that already incorporate structural meaningful rules, thus providing a partial solution from the outset. Moreover, the model reasoning capabilities, in conjunction with performance feedback and sensory-motor data, allow for targeted and efficient policy refinements. 

This approach offers a distinct advantage over traditional model-free learning methods, such as reinforcement learning, which typically start from random control strategies and rely on stochastic exploration to achieve improvement.

\subsection*{3.2 Control problems}

We assessed the effectiveness of the proposed method using a suite of classic control tasks from Gymnasium (previously known as Gym~\cite {gym:24}). In addition to the five standard Gymnasium benchmark tasks-Acrobat, CartPole, MountainCar, MountainCarContinuous, and Pendulum- we also included the InvertedPendulum task from the MuJoCo library. \ma{We deliberately focused on these traditional, well-understood environments because our goal is to assess a novel capability: whether LLMs can directly generate functional controllers for embodied agents and whether Iterative Policy Refinement can reliably improve such controllers. These relatively simple benchmarks provide a clear and interpretable baseline for establishing viability before adressing more complex control scenarios. We emphasize that, although these tasks lie at the simpler end of standard continuous-control benchmarks, they nonetheless capture several fundamental challenges of sensory–motor control, including partial observability, delayed reward, nonlinear dynamics, and stabilization under perturbations. As such, they are well suited for validating the core methodological contribution of this work.}

\ma{The choice of environments was also motivated by practical considerations related to computational cost. Evaluating an LLM-based policy generation framework is substantially more expensive than standard learning loops, as it requires repeated environment rollouts, controller execution, and multiple cycles of LLM prompting and refinement. Conducting the large-scale experimental sweep needed for a robust assessment of Iterative Policy Refinement therefore necessitated environments with short episode lengths and fast simulation times. While success on these classic control tasks establishes an essential baseline, it does not by itself characterize the limits of LLM-based control in higher-dimensional, contact-rich domains such as the Humanoid task. Extending the proposed approach to these more challenging environments is a natural next step, but doing so will require additional engineering and computational resources—such as longer-horizon evaluations, more expensive simulations, and potentially richer controller parameterizations—which we discuss in Section 5.
}

To ensure that the solutions generated by LLMs are not directly reproduced from memorized training data, we introduced two modified versions of the CartPole task: (i) Cartpole *1, which varies the encoding of the action space, and (ii) Cartpole *2, which alters the encoding of both the observation and action spaces. A detailed description of all tasks is provided in Appendix B. As we will demonstrate, the generation of original solutions is further supported by the analysis of the solution-generation process. 

\subsection*{3.3 Experimental conditions}

\m{To evaluate the effectiveness of the proposed method, we conducted a comprehensive series of experimental runs, systematically varying the LLM, the temperature parameter—which controls the stochasticity in next-token prediction—and the control task. Each experimental condition was replicated ten times to ensure statistical robustness and allow for meaningful comparisons across models and settings.}

The evaluated models included \textbf{LLaMa3.3} (70 billion parameters)~\cite{llama}, \textbf{Mistral-Large} (123 billion parameters)~\cite{mistral}, \textbf{Qwen2.5} (72 billion parameters)~\cite{qwen}, \textbf{DeepSeek-R1} (70 billion parameters)~~\cite{deepseek} and \textbf{GPT-oss:120b} (120 billion parameters)~~\cite{gpt-oss}.\m{We also ran experiments with the GPT-oss:20b (20 billion parameters, which yielded markedly worse performance than the larger model.} All models employed k-quantization techniques for weights quantization~\cite{kim:23}, specifically the Q4\_K\_M variant. The models were hosted locally using an Ollama server\footnote{https://ollama.com/} and executed on an NVIDIA DGX cluster with eight H100 GPUs.

\m{Several methodological and practical considerations guided the selection of these specific models. First, all chosen models are open-source, ensuring complete transparency of the architectures and inference procedures and enabling local deployment without reliance on commercial APIs. This was a critical requirement for our experimental framework, which involved repeated, fine-grained evaluations under varying control conditions and the collection of sensory-motor interaction data. The open-source nature of these models also facilitated consistent experimental control and reproducibility across replications.}

\m{Second, we deliberately selected relatively compact models—ranging from 70 to 123 billion parameters—rather than the largest available LLMs. This choice reflects a trade-off between representational power and computational cost. While larger models may offer enhanced reasoning or generalization abilities, their high resource demands would have severely limited the number of experimental replications and parameter combinations that could be feasibly tested. By employing smaller yet high-performing models, we were able to perform multiple replications (ten per condition) and systematically investigate the impact of various temperature settings and task characteristics on model performance. This design choice enabled a more rigorous statistical assessment of the models’ behavior and of the stability of the learning process.}

\m{Moreover, these models represent state-of-the-art architectures within their respective families, combining efficient transformer backbones with advanced training methodologies such as mixture-of-experts routing, instruction tuning, and extended context handling. As such, they provide a representative spectrum of contemporary open-weight LLM designs, allowing us to examine whether the observed control capabilities are general across architectures or depend on specific model properties.}

\m{For the Cartpole*1 and Cartpole*2 tasks, we varied the model temperature across the set of [$0.0, 0.4, 0.8, 1.6$ and $3.2$], thereby assessing the influence of stochasticity on the exploration–exploitation balance during policy generation. For all other control tasks, we limited models and temperature to the two setups that yielded the best results in preliminary experiments on the CartPole variants. The iterative learning process was run for 100 epochs for the CartPole tasks and 500 epochs for the InvertedPendulum task. The performance of the resulting control policies was evaluated by averaging the rewards obtained over 20 evaluation episodes, each initialized with randomly varied initial conditions. For the remaining tasks, training was limited to 50 epochs, and performance was assessed over 10 episodes.}

\m{This experimental design thus ensured a balanced combination of computational tractability, statistical rigor, and architectural diversity, providing a robust empirical foundation for evaluating the proposed LLM-based control methodology.}

\subsection*{3.4 Evaluation metrics}

\m{Model performance in embodied control tasks spans multiple complementary dimensions that cannot be adequately captured by a single metric. Specifically, it is necessary to consider the proportion of experimental runs that successfully reach an optimal solution (success rate), the stability of the obtained controllers when evaluated across different episodes and random initializations (robustness), and the efficiency with which solutions are discovered during the iterative learning process (learning time). Because performance along these dimensions can vary independently, a model may converge rapidly but produce unstable controllers, or consistently identify solutions that lack robustness—no single metric alone can adequately identify the "best" model. We therefore designed a composite Figure of Merit (FoM) that integrates these complementary measures, enabling fair model-level comparisons and clear rankings across experimental conditions. This aggregate score serves solely as a concise summary for comparative purposes. Throughout our analysis, we also report the underlying metrics separately, allowing readers to examine each performance dimension in isolation and draw their own conclusions about model behavior.}

\m{To post-evaluate the results of the experiments, we designed the following evaluation metrics:}

$Average\ Reward$: Quantifies the mean reward achieved by each tested policy throughout the iterated learning process across episodes and replications. 

$Success$: Denotes the proportion of experimental replications in which the model successfully identified an optimal solution.

$Learning\ Time$: Evaluates the time taken for finding the optimal solution during the iterated learning process.

$Robustness$: Assesses the capability of the best policy to handle varying environmental conditions.

$Figure\ of\ Merit (FoM)$: A composite index that provides an overall performance evaluation by aggregating  Success, Robustness, and Learning Time. 

The $Average\ Reward$ of a model is computed by averaging the rewards obtained across all learning episodes and replications, as defined by the following equation:

\begin{equation}
    \overline{R} = \frac{1}{N_{\text{reps}} \cdot N_{\text{episodes}}} \sum_{j=1}^{N_{\text{reps}}} \sum_{k=1}^{N_{\text{episodes}}} R^{(j,k)}
\end{equation}
where:
\begin{itemize}
    \item \( N_{\text{reps}} = 10 \): Number of replications,
    \item \( N_{\text{episodes}} = 100 \): Number of learning episodes,
    \item \( R^{(j,k)} \): Reward for replication $j$ and learning episode $k$.
\end{itemize}

The $Success$ of a model is determined using the following expression:

    \textbf{Success} ($Q$): Measures success frequency across replications:
    \begin{equation}
        Q_{\text{success}} = \frac{1}{N_{\text{reps}}} \sum_{j=1}^{N_{\text{reps}}} \mathbb{I}\left(\exists\, t \leq T_{\max}: R^{(j,t)} = R_{\max}\right)
    \end{equation}
    where $N_{\text{reps}} = 10$ independent replications, $T_{\max} = 100$ episodes, and $R_{\max} = 500$ is the maximum reward. Range: $[0, 1]$.

The $Learning\ Time$ and $Robustness$ metrics are calculated only for those replications in which the model successfully identified an optimal solution. These are defined by the following equations:

\begin{enumerate}
    \item \textbf{Robustness} ($\overline{Q}_{\text{robustness}}$): For the first strategy that achieved maximum reward in each of the $N_{reps}$ replications, measures strategy robustness across initial conditions:
    \begin{equation}
        \overline{Q}_{\text{robustness}} = \frac{1}{|\mathcal{S}_m|} \sum_{j \in \mathcal{S}_m}  \frac{1}{N_{\text{episodes}}} \sum_{k=1}^{N_{\text{trials}}} \mathbb{I}\left(R^{(j,k)} = R_{\max}\right)
    \end{equation}
    where $N_{\text{episodes}} = 2000$, $R^{(j,k)}$ is the reward for the learning episode $k$ in replication $j$, and $R_{\max} = 500$ is the maximum reward. Range: $[0, 1]$.

    \item \textbf{Learning Time} ($\overline{Q}_{\text{time}}$): For the first strategy that achieved maximum reward in each of the $N_{reps}$ replications, measures convergence rapidity as the fraction of episodes needed to achieve the maximum reward:
    \begin{equation}
        \overline{Q}_{\text{time}} = \frac{1}{|\mathcal{S}_m|} \sum_{j \in \mathcal{S}_m} \frac{\min\{t \in [1, T_{\max}] : R^{(j,t)} = R_{\max}\}}{T_{\max}}
    \end{equation}
    where \( T_{\max} = 100 \) is the maximum number of episodes, and $R_{\max} = 500$. Range: \([0.01, 1]\).
\end{enumerate}

For both $Robustness$ and $Learning\ Time$, $\mathcal{S}_m$ denotes the subset of successful replications in which the model achieved the maximum reward at least once:
\[
\mathcal{S}_m = \left\{j \in \{1,\ldots,N_{\text{reps}}\} \, \bigg| \, \exists\, t \leq T_{\max}: R^{(j,t)} = R_{\max}\right\}
\]

The $Figure\ of\ Merit$ for a model is computed using the following formulation:

\begin{equation}
    \text{FoM} = \frac{\overline{Q}_{\text{stability}} \cdot \left(Q_{\text{success}}\right)^2}{\overline{Q}_{\text{time}}}
\end{equation}

To emphasize the importance of consistently discovering optimal solutions, the Success component is squared. This design choice penalizes models that fail to solve one or more replications, even if they exhibit high stability or fast convergence in other replications.

\subsection*{Results}

Figure 2 and 3 present the results obtained for the two versions of the CartPole task under varying temperature settings. In Figure 2, performance is shown across increasing temperature values selected from the range [0.0, 0.4, 0.8, 1.6, and 3.2].  Figure 3 displays performance at the optimal temperature level for each model. The GPT-oss:120b model achieves the highest performance, obtaining statistically superior results relative to all other models. In comparison, the Qwen2.5 and Llama3.3:70b models yield substantially lower performance, while Mistral-large123b and DeepSeek-r1:70b perform markedly worse and show no significant difference from one another.

Statistical significance was assessed using the Shapiro-Wilk test for normality, followed by a Kruskal-Wallis non-parametric ANOVA for overall group comparison. Post-hoc pairwise differences were evaluated using Dunn’s test.

\m{Figure 4 provides a detailed comparison of model performance across the two CartPole tasks at the optimal temperature setting. The GPT-oss:120b model significantly outperforms the other models with respect to the composite Figure of Merit evaluation metric (left panels). As shown in the right panels, this advantage is primarily attributable to its learning time, which is considerably faster than that of the other models. Moreover, this superiority stems from the success rate (the proportion of replications in which the model successfully discovers optimal solutions across both tasks). Qwen2.5:72b is the second-best performer with respect to the Figure of Merit. Its relative advantage over the other models arises from faster learning on the CartPole*1 task and a higher success rate on the CartPole*2 task.}

The relatively poor performance of DeepSeek-R1, despite its strong capabilities in traditional reasoning tasks and in mathematical problems, suggests that embodied control and iterated learning rely on distinct forms of reasoning not captured by conventional benchmarks.

The control strategies proposed by the LLMs are the result of a dynamic reasoning process that combines domain-specific knowledge with iterative refinement. Consequently, they are not simply derived from solutions experienced in the training data. Evidence for this can be seen in the low initial performance of the Qwen2.5-72B model during the first iteration (see Table 1), where the maximum reward is 44.1 and 49.85 for the CartPole*1 and CartPole*2 tasks, respectively. Performance improves significantly over subsequent iterations, reaching optimal rewards in 7 out of 10 replications in the CartPole*1 and in all 10 replications for CartPole*2. The three replications that did not reach optimal performance yielded maximum rewards of 66.85, 275.95, and 410.25, respectively (Table 1). \m{The gpt-oss:120b model manages to generate optimal policies during the first iteration. However, this happens only in the case of the Cartpole*1 task and only in 4 out of 10 replications.} These findings are further supported by the experimental design, which intentionally deviates from standard benchmark formulations to prevent the eventual usage of memorized solutions. The models thus engage in genuine problem-solving. \ma{As a point of reference, modern RL methods also achieve maximum performance on the CartPole benchmark~\cite{shankar2025policy,asri2024physics,openai_gym_wiki_leaderboard}.}

As discussed, the optimal value of the temperature hyperparameter is both model- and task-dependent. \m{However, to keep the computational cost feasible, we conducted additional experiments on the remaining tasks by testing the two best models in the CartPole tasks and the two best temperatures for each of them. The models tested were GPT-oss and Qwen2.5 and the best temperatures were, respectively, [$0.0$, $0.8$], and [$0.0$, $1.6$]. As in previous tests, each experimental condition was replicated 10 times}. 

Table 2 summarizes the results of these additional tasks. These problems do not have a predefined optimal reward. In any case, the GPT-oss model achieves by far the highest performance on four of the five tasks, while Qwen2.5 has the highest performance on the Inverted Pendulum, closely followed by GPT-oss. A notable case is the Pendulum task, where Qwen2.5's performance is notably poor compared to the high performance of GPT-oss. As with the CartPole tasks, initial performance in the first iteration is limited, underscoring the crucial role played by iterative learning and ensuring that the models do not simply retrieve predefined solutions available in the data used for the LLMs training process. \ma{As a point of reference, modern RL methods~\cite{asri2024physics,chou2017improving,openai_gym_wiki_leaderboard} typically achieve average returns of approximately -263 for Pendulum, -138 for Acrobot, 950 for Inverted Pendulum, -110 and 90 on the discrete and continuous variants of MountainCar, respectively.}

The solution process followed by the models can be illustrated through an analysis of the best-performing CartPole*2 replication carried out with  Qwen2.5 (see also Appendix C). The complete experimental data, including logs, control rules, and videos, are available in the online repository: https://jtyska.github.io/llm-robotics-article/. 

The initial control strategy generated by the model, based solely on a textual description of the environment, agent, and task goal, consists of a structured set of rules. These rules attempt to correctly determine the agent’s actions by considering the following states: (i) the direction in which the pole is leaning, (ii) the proximity of the cart to the left or right boundary, (iii) whether the pole’s angular velocity exceeds certain thresholds, (iv) whether the cart’s linear velocity exceeds certain thresholds, (v) whether the pole is (nearly) vertical.

Although the initial strategy includes a comprehensive set of plausible control rules, it fails to correctly prioritize the most critical factors and is structured in a way that only the rules based on the pole’s angle are actually executed, while the remaining rules remain inactive. Consequently, the initial average reward is limited to 49.85 out of a maximum of 500.

Through iterative learning, the strategy is incrementally refined. In subsequent iterations, the average reward increases to 116.1 and 296.55, eventually reaching the maximum reward of 500 in the fourth iteration. These improvements arise by determining the action based on both the angular velocity of the pole and the angle of the pole, and by prioritizing the importance of the former. The model proactively analyzes the sensory-motor feedback from each iteration, identifies limitations, and proposes targeted modifications to enhance performance.

Despite these advancements, the models exhibit occasional errors and hallucinations, consistent with findings in prior work (e.g., Li et al., 2024~\cite{li2024embodied}; Wang et al., 2024~\cite{wang24planning}). For example, the model may misinterpret one sensor reading as another during introspective analysis. Such errors are observed in all replications but do not prevent the models from eventually discovering optimal control policies.

\m{To assess the impact of the amount of sensory-motor data processed by the model during each iteration, we conducted a series of control experiments in which the number of sensory-motor steps was set to 5, 20, 50, and 100. The analysis was performed using the gpt-oss:20b model to limit the computational cost and on the Pendulum Task, i.e., the task LLMs for sensory-motor control that resulted in more difficulty. The results, presented in Figure 5, indicate that both too few and too many sensory-motor steps lead to reduced performance, with optimal performance achieved at an intermediate number of steps. A comparison of results from the gpt-oss:20b and gpt-oss:120b models using 20 sensory-motor steps, also presented in the figure, confirms that, as expected, larger models deliver superior performance.
}

\m{To evaluate the effect of prompt accuracy, we performed a control experiment on the Pendulum task in which the task description was made more precise. In particular, we explicitly described how the sign of the action value determines the direction (clockwise or counter-clockwise) of the torque applied to the pendulum. As expected, the experiment using the more accurate prompt produced substantially better results. These findings are illustrated in the first two histograms of Figure 6. Providing a fully detailed description of the agent, its sensory-motor system, and the environment may be infeasible, especially in settings involving complex agents or environments. The model’s ability to generate effective solutions even when provided with incomplete or partially inaccurate descriptions therefore represents an important strength of the proposed approach.}

\m{Finally, to assess the contribution of different key components of the method, we carried out a series of ablation experiments in which one or more elements were removed from Prompt 4, which instructs the model to generate an improved control strategy. As shown in Figure 6, performance deteriorates drastically in the NoData condition, where all information from previous attempts and strategies is removed (i.e., the current, previous, and best strategies with their associated rewards, as well as the sensory-motor data collected during the evaluation of the current strategy). Removing only the sensory-motor data (NoSensoryMotorData), the previous strategy with its reward (NoPreviousStrategy), or the best strategy with its reward (NoBestStrategy) leads to a moderate reduction in performance. All conditions differ statistically from the baseline (MoreAccuratePrompt), except for the case where the previous strategy was not provided (NoPreviousStrategy). We assessed distributional normality using the Shapiro–Wilk test and subsequently applied the Mann–Whitney U test to evaluate statistical significance, as the data did not meet the assumption of normality.}

\begin{figure}[!htbp]
\centering
\includegraphics[width=\linewidth]{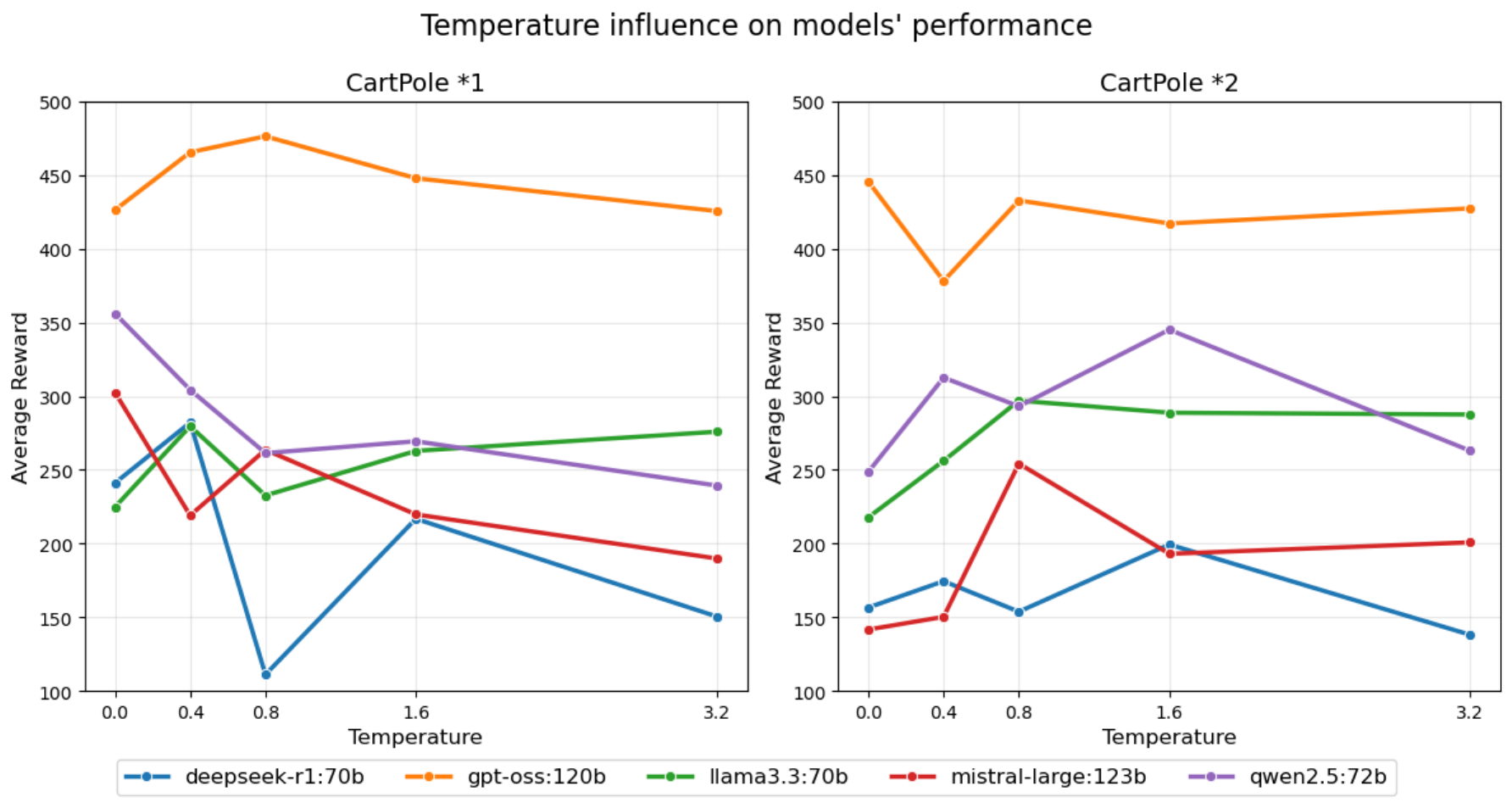}
\caption{ Average reward obtained for the CartPole1 (left) and CartPole2 (right) tasks. The horizontal axis represents the temperature value, while the vertical axis shows the average reward, computed over 10 replications. Line colors indicate the different models evaluated.}
\label{fig:cartpole_m1_temp}
\end{figure}

\begin{figure}[!htbp]
\centering
\includegraphics[width=\linewidth]{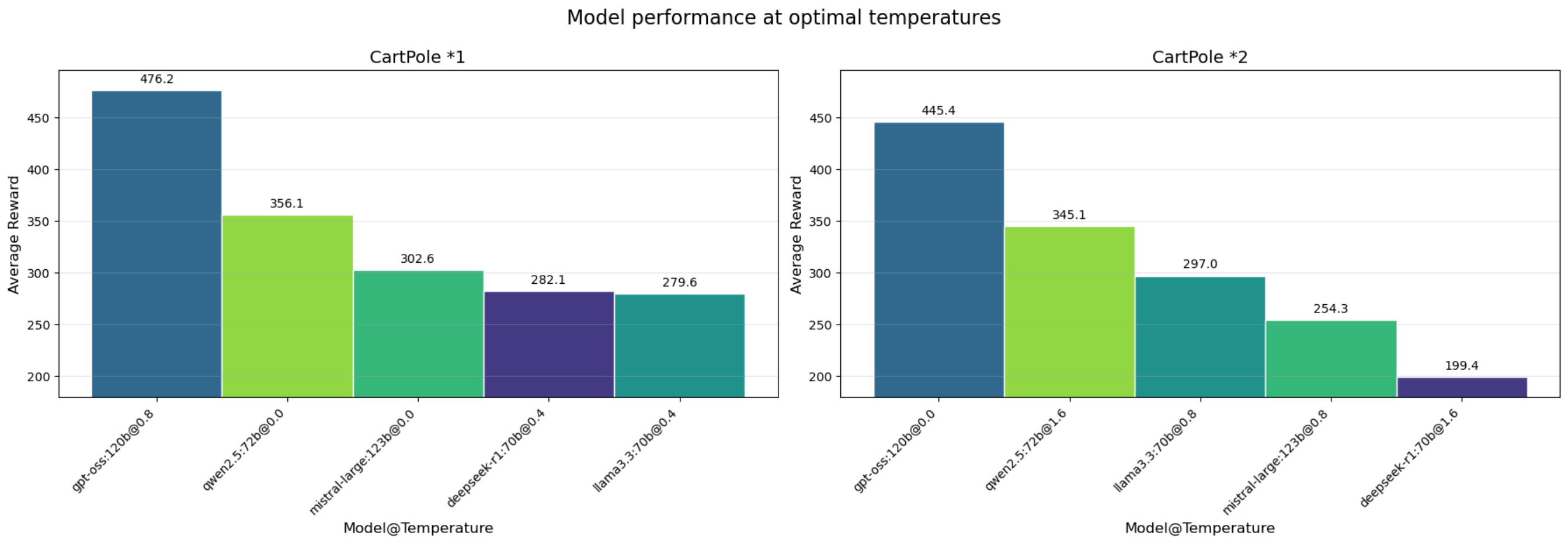}
\caption{Average reward at the optimal temperature setting for the Cartpole*1 and Cartpole*2 tasks (left and right, respectively). The vertical axis represents the average reward, while the horizontal axis specifies the LLM used.}
\label{fig:cartpole_best_performance}
\end{figure}

\begin{figure}[!htbp]
\centering
\includegraphics[width=\linewidth]{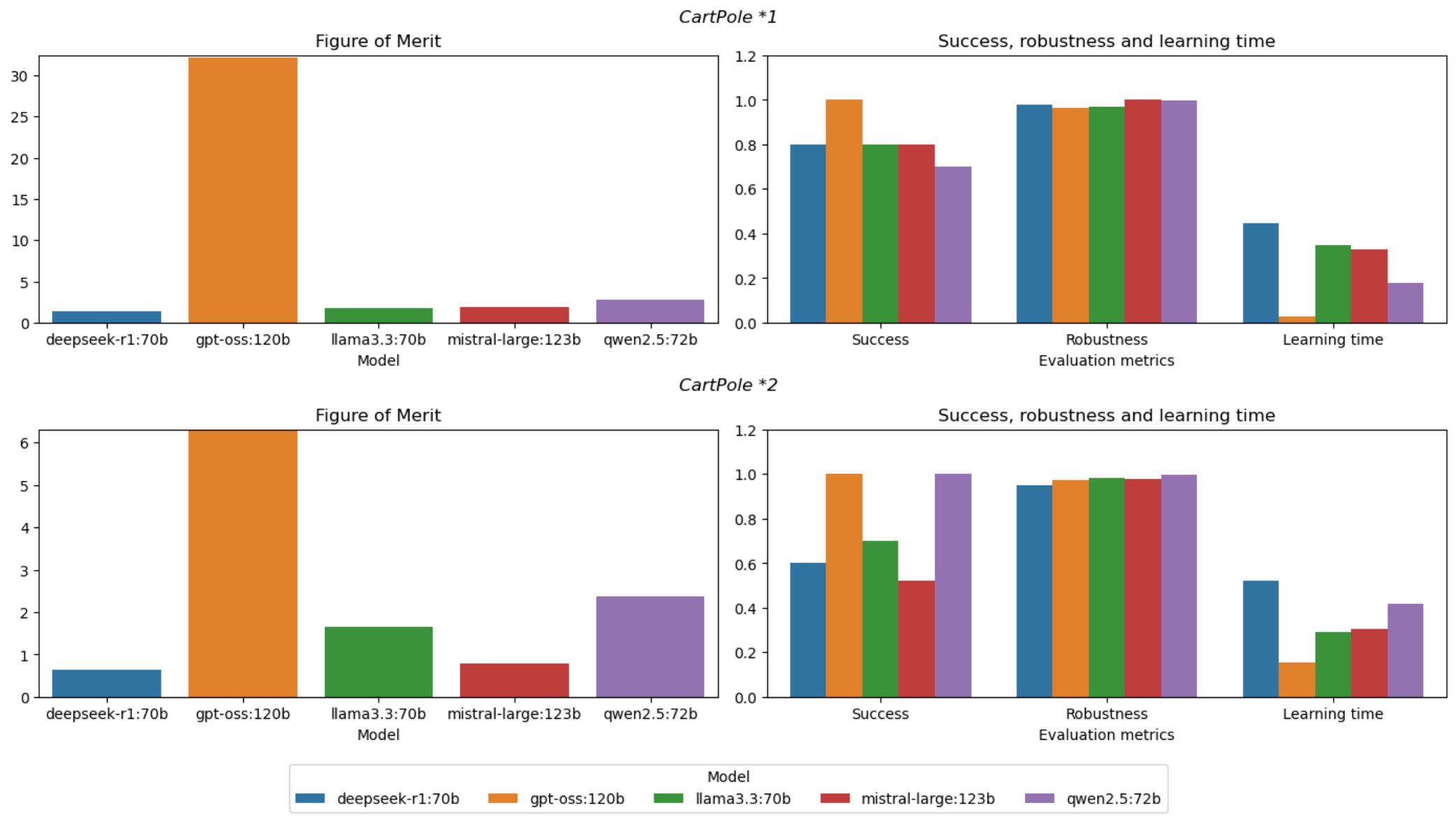}
\caption{\m{Performance metrics for the CartPole *1 (top) and CartPole *2 (bottom) tasks. The plots on the left display the Figure of Merit, while the plots on the right show Success, Robustness, and Learning Time. Histogram colors indicate the LLM used. Results are reported for the optimal temperature setting. Note that the y-axis scales differ between the left-hand plots. 
}}
\label{fig:cartpole_detailed_performance}
\end{figure}


\begin{table}[]
\caption{\m{Rewards obtained during the first iteration and during the iteration that achieved the highest reward. For both the first and best iterations, the table reports the average rewards across all replications, as well as the rewards achieved in the best and worst replications. The results were obtained on the CartPole*1 and CartPole*2 tasks using the deepseek-r1:70b, llama3.3:70b, mistral-large:123b, qwen2.5:72b, and GPT-oss:120b models with their optimal temperature settings. Reward values range from 0 to 500, with higher values indicating better performance.}}

\label{tab:cartpole_performance}
\begin{tabular}{llccc|ccc}
\hline
\multicolumn{1}{l}{}         & \multicolumn{1}{l}{} & \multicolumn{3}{c|}{\textbf{First iteration reward}}                                                                                                           & \multicolumn{3}{c}{\textbf{Best iteration reward}}                                                                                                                                \\ \hline
\textbf{Environment}         & \textbf{Model}       & \textbf{\begin{tabular}[c]{@{}c@{}}Worst\\ replication\end{tabular}} & \textbf{\begin{tabular}[c]{@{}c@{}}Best\\ replication\end{tabular}} & \textbf{Average} & \textbf{\begin{tabular}[c]{@{}c@{}}Worst\\ replication\end{tabular}} & \textbf{\begin{tabular}[c]{@{}c@{}}Best\\ replication\end{tabular}} & \multicolumn{1}{l}{\textbf{Average}} \\ \hline
\multirow{4}{*}{CartPole *1} & deepseek-r1:70b      & 8.70                                                                 & 45.75                                                               & 12.60            & 473.75                                                               & 500                                                                 & 495.78                               \\
                             & llama3.3:70b         & 9.20                                                                 & 172.65                                                              & 74.45            & 352.35                                                               & 500                                                                 & 484.46                               \\
                             & mistral-large:123b   & 8.60                                                                 & 8.85                                                                & 8.78             & 52.90                                                                & 500                                                                 & 429.50                               \\
                             & qwen2.5:72b          & 41.55                                                                & 44.10                                                               & 43.05            & 66.85                                                                & 500                                                                 & 425.31                               \\
 & gpt-oss:120b & 42.15& 500.00& 344.32& 500.00& 500&500.00\\ \hline
\multirow{4}{*}{CartPole *2} & deepseek-r1:70b      & 9.60                                                                 & 40.15                                                               & 18.57            & 376.65                                                               & 500                                                                 & 472.46                               \\
                             & llama3.3:70b         & 8.80                                                                 & 315.05                                                              & 71.24            & 470.30                                                                & 500                                                                 & 495.80                               \\
                             & mistral-large:123b   & 9.00                                                                 & 190.80                                                              & 41.54            & 124.95                                                               & 500                                                                 & 415.19                               \\
 & qwen2.5:72b          & 8.70                                                                 & 49.85                                                               & 18.31            & 500.00& 500                                                                 &500.00                               \\
                             & gpt-oss:120b& 9.8& 486.1& 245.99& 500.00& 500& 500.00\\ \hline
\end{tabular}
\end{table}

\begin{table}[ht]
\centering
\renewcommand{\arraystretch}{1.2}
\begin{tabular}{l l c c c c c}
\hline
Environment & Model & \makecell{Optimal \\ Temperature} & \makecell{Mean \\ Learning Time} & \makecell{Mean \\ Best Reward} & \makecell{Mean \\ Initial Reward} & \makecell{Mean \\ Improv. Ratio} \\
\hline
\multirow{2}{*}{Acrobot} 
 & \textbf{gpt-oss} & 0.8 & 0.03 & \textbf{-77.30} & -500.00 & 0.99 \\
 & qwen2.5 & 1.6 & 0.04 & -108.24 & -500.00 & 0.92 \\
[0.8em] 
\multirow{2}{*}{Inverted Pendulum} 
 & gpt-oss & 0.0 & 0.31 & 738.58 & 12.45 & 0.73 \\
 & \textbf{qwen2.5} & 1.6 & 0.19 & \textbf{829.66} & 5.05 & 0.83 \\
[0.8em] 
\multirow{2}{*}{\makecell{MountainCar\\Continuous}} 
 & \textbf{gpt-oss} & 0.8 & 0.04 & \textbf{94.81} & -99.90 & 0.99 \\
 & qwen2.5 & 0.0 & 0.04 & 62.22 & -56.26 & 0.78 \\
[0.8em] 
\multirow{2}{*}{\makecell{MountainCar\\Discrete}} 
 & \textbf{gpt-oss} & 0.8 & 0.16 & \textbf{-111.54} & -200.00 & 0.87 \\
 & qwen2.5 & 1.6 & 0.07 & -166.24 & -200.00 & 0.33 \\
[0.8em] 
\multirow{2}{*}{Pendulum} 
 & \textbf{gpt-oss} & 0.8 & 0.18 & \textbf{-190.31} & -1725.44 & 0.96 \\
 & qwen2.5 & 0.0 & 0.13 & -911.81 & -1334.76 & 0.35 \\
\hline
\end{tabular}
\caption{\m{Performance results for the Acrobot, Inverted Pendulum, MountainCar Continuous, MountainCar Discrete, and Pendulum tasks obtained using GPT-oss and Qwen2.5 models with optimal temperature settings. For each task, the table reports the optimal temperature, mean learning time, mean best reward, mean initial reward, and mean improvement ratio. “Mean” indicates that these four values are averaged across replications. The improvement ratio is the percentage gain over the initial performance, measured against the task’s maximum reward (when defined) or otherwise the best reward achieved for that task across all experiments. In all tasks, performance improves with higher values.}}
\label{tab:results}
\end{table}

\section*{Discussion}

We presented a method that enables large language models (LLMs) to control embodied agents, allowing them to exhibit goal-directed behavior. Unlike approaches that rely on predefined motor primitives, \m{our method generates control policies} that directly map continuous observation vectors to continuous action vectors, thereby controlling the agent's actuators directly. Moreover, the method operates without requiring model fine-tuning, thereby eliminating the need to collect large-scale sensory-motor datasets.

The proposed approach leverages the in-context learning capabilities of LLMs in conjunction with an iterative learning process. \m{Initially, the model is prompted to generate a control strategy based on a verbal description of the agent, the environment, and the intended goal. This strategy is subsequently refined through repeated prompting, in which the model is asked to reflect on the performance of the current (and eventually previous) strategy using sensory-motor data obtained during evaluation, and to revise the strategy accordingly. This iterative refinement mechanism, supported by performance feedback and sensory-motor data, enables the model to identify and eliminate maladaptive control rules that may emerge from reasoning errors, hallucinations, or the inherent incompleteness of the problem description. We validated the proposed method using five relatively compact LLMs on a suite of classic control tasks from the Gymnasium library, as well as the inverted pendulum task from the MuJoCo library.}

\m{The results obtained in this study highlight the distinctive potential of large language models (LLMs) as controllers for embodied and situated agents. Unlike conventional approaches that rely on extensive training data or explicit demonstrations, LLMs can synthesize plausible initial control strategies on the fly based solely on a verbal description of the agent, its environment, and the task to be achieved. This ability stems from their vast knowledge and abstract reasoning skills, which allow them to infer the structure of plausible, effective control rules. Moreover, LLMs can serve as reasoning engines capable of hypothesizing, testing, and refining behavioral strategies through iterative interaction.
}

\begin{figure}[!htbp]
\centering
\includegraphics[width=\linewidth]{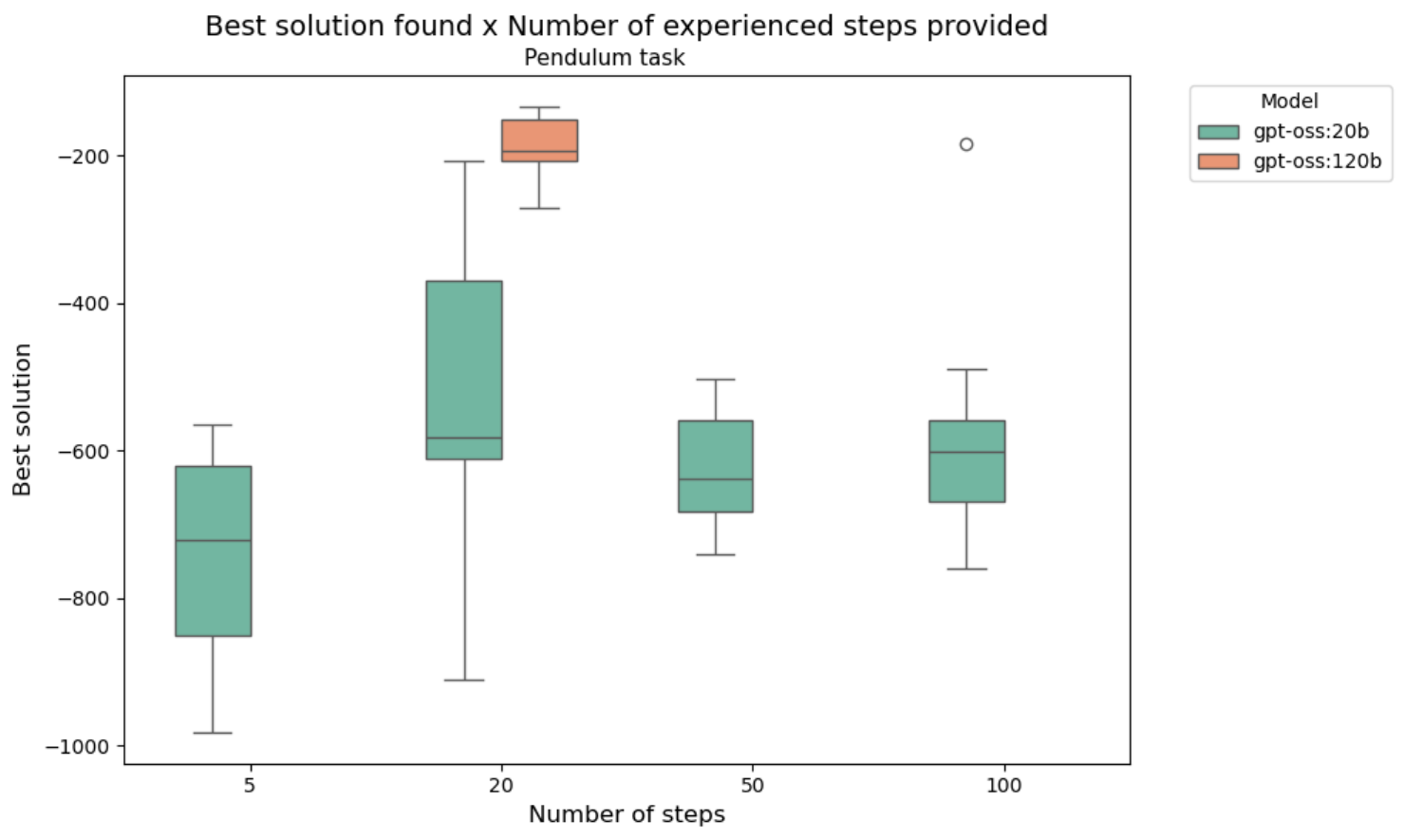}
\caption{\m{Experiments conducted by varying the amount of sensory-motor data provided to the model. All experiments were performed on the Pendulum task. The green and orange boxplots correspond to the gpt-oss:20b and gpt-oss:120b models, respectively. Each boxplot summarizes the results of 10 replications. }}
\label{fig:number_steps_analysis}
\end{figure}

\begin{figure}[!htbp]
\centering
\includegraphics[width=\linewidth]{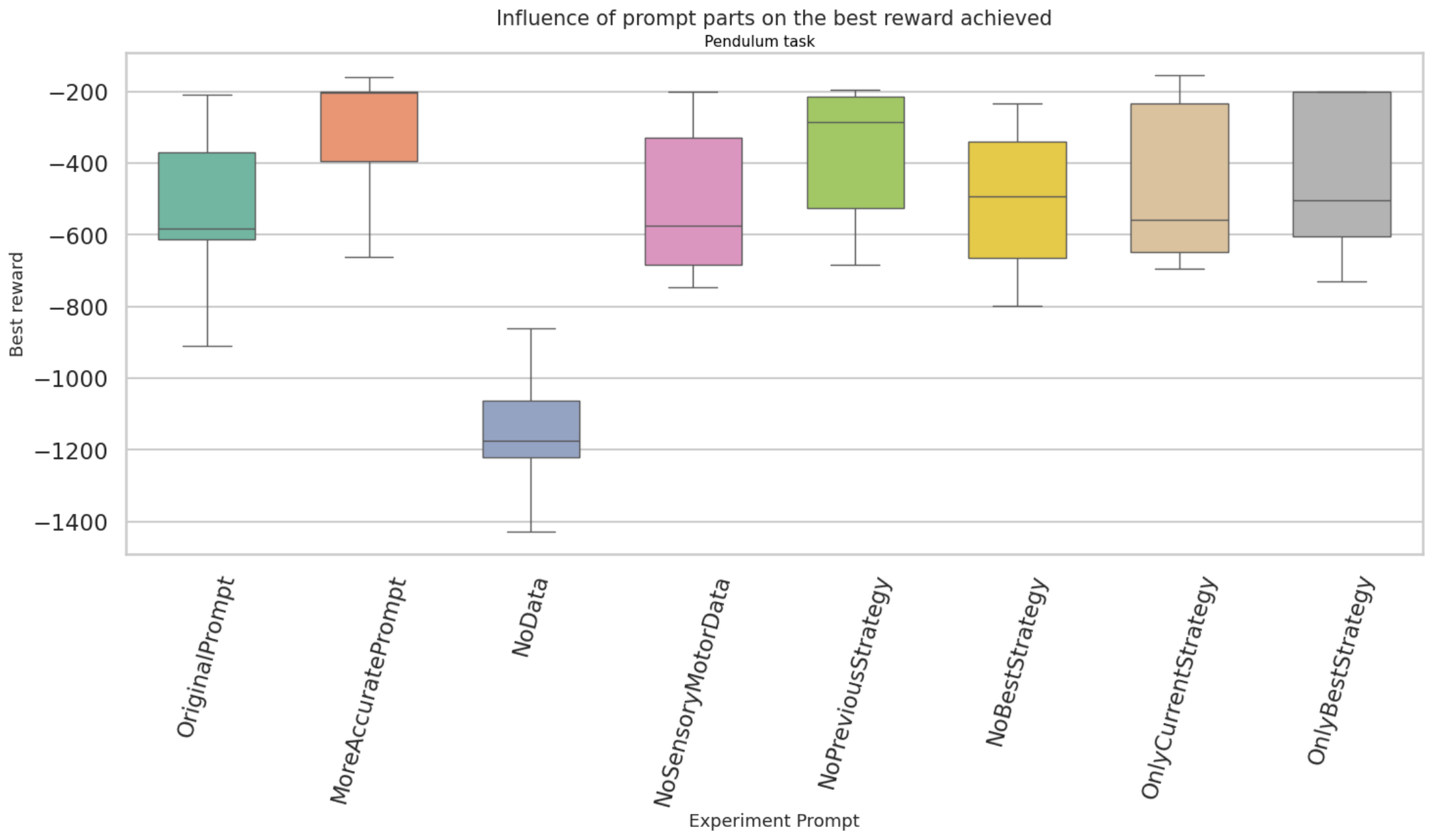}
\caption{\m{Experiments conducted by varying the accuracy of the task description in the prompt and by ablating model components. All data were collected on the Pendulum task using the gpt-oss:20b model. \textbf{OriginalPrompt}: baseline task description used in the main experiments; \textbf{MoreAccuratePrompt}: revised, more precise task description with ambiguities removed and additional task-relevant details; \textbf{NoData}: Prompt 4 is omitted entirely (no current and prior strategies, rewards, or sensory-motor data); \textbf{NoSensoryMotorData}: sensory-motor data are omitted; \textbf{NoPreviousStrategy}: the previous strategy with its associated reward is omitted; \textbf{NoBestStrategy}: the best strategy with its associated reward is omitted; \textbf{OnlyCurrentStrategy}: only the current strategy with its associated reward and sensory-motor data is included; \textbf{OnlyBestStrategyData}: only the best strategy with its associated reward is included. 
}}
\label{fig:prompt_influence}
\end{figure}

\m{A central advantage of this paradigm is that it enables the rapid initialization of goal-directed behavior. When prompted with a textual specification, LLMs can draw on prior conceptual and procedural knowledge to propose a structured control policy that already encodes meaningful relationships between sensory and motor states. This contrasts sharply with reinforcement learning (RL) or other model-free methods, which typically begin from random policies and require extensive exploration to converge on a viable solution. By starting from an informed, reasoned hypothesis, LLM-based agents can bypass the most computationally expensive early phases of exploration, thus accelerating the learning process and reducing the amount of interaction data needed to reach optimal or near-optimal performance.
}

\m{Moreover, the reasoning abilities of LLMs make it possible to analyze and interpret the outcomes of previous control attempts in a qualitatively different manner from statistical learners. During iterative refinement, the model can explicitly describe and justify modifications to its own strategy—inferring, for instance, which variables have the greatest influence on success or which control rules should be prioritized or suppressed. This metacognitive capability allows LLMs to apply targeted changes that improve behavioral efficacy without requiring large numbers of stochastic trials. The ability to combine explicit reasoning about system dynamics with empirical feedback from sensory-motor data provides a powerful form of hybrid learning that integrates symbolic and sub-symbolic knowledge sources.}

\m{From a broader perspective, this integration suggests that LLM-based control may substantially reduce the dependency on manually curated datasets and physical-world demonstrations that currently constrain data-driven robotics. Instead, control strategies can be generated, evaluated, and iteratively refined, guided by the LLM’s internal knowledge of physical regularities, task structures, and causal dependencies. As demonstrated in our experiments, even relatively compact models such as GPT-oss:120b and Qwen2.5:72b were able to infer missing parameters and converge toward optimal control solutions. This indicates that the knowledge encoded in modern language models can serve as a versatile prior for embodied learning, bridging the gap between high-level reasoning and continuous motor control.}

\m{Despite these promising results, several limitations remain that constrain the scalability and practical deployment of the proposed approach. First, the computational cost associated with employing very large models—often comprising tens or hundreds of billions of parameters—represents a significant bottleneck. Although our experiments show that relatively compact models such as GPT-oss:120b and Qwen2.5:72b already achieve competitive performance, real-world applications requiring continuous and high-frequency control would demand even greater computational efficiency. Future research should therefore investigate strategies for model distillation, parameter-efficient fine-tuning, or modular architectures that allow smaller specialized models to handle distinct aspects of embodied control while retaining access to shared linguistic and reasoning capabilities.}

\m{A second limitation concerns the intrinsic restriction imposed by the finite context window of current LLMs. The number of input tokens that can be processed at once limits both the amount of sensory-motor data collected during policy evaluation and the quantity of performance feedback that can be incorporated during iterative learning. In our experiments, including only 20 sensory-motor steps yielded the best results. However, more complex problems and larger LLMs may require substantially larger amounts of data. Expanding the effective memory span of the models or developing hierarchical compression mechanisms for sensory-motor sequences will be essential for extending the applicability of LLM-based control to more complex, long-horizon tasks.
}

\ma{Building on these considerations, a deeper investigation of scalability is crucial. As task complexity increases, particularly in the form of higher-dimensional state and action spaces, longer planning horizons, or partially observable environments, the computational and contextual costs related to the use of LLM grow substantially. Larger observation and action spaces require more extensive sensory–motor representations, thereby expanding the size of the input prompt and increasing the token budget needed to encode both the policy and its historical context. Similarly, long-horizon tasks necessitate broader windows of past experience to capture temporal dependencies, goal structure, or delayed effects. While including more interaction history can improve planning, it also raises the risk of overwhelming the model with low-level numerical data, which may distract from the high-level task objective and degrade reasoning quality.}

\ma{A key mechanism for addressing these challenges is the Iterative Policy Refinement framework itself. By actively exploring and reflecting over successive policy evaluations, the LLM agent can gradually discover and incorporate missing information, allowing it to cope with initially incomplete state representations or partial observability. This iterative design effectively enables the agent to “fill in the gaps” of its world model through structured trial-and-error. To further support scaling to more complex control problems, recent developments in agentic AI, such as the Model Context Protocol, offer promising architectural solutions. These protocols provide standardized methods for dynamically managing large and structured context, enabling the integration of modular memory, subtask decomposition, and feedback-guided control. Incorporating such capabilities could systematically mitigate the scalability limits identified here and support LLM-based agents operating in high-dimensional, sequential, and partially observable domains.}

\m{A third limitation involves the current dependency on manually constructed prompts. Although structured prompting proved crucial for eliciting coherent control strategies and guiding iterative refinement, it still requires significant human expertise to design effective prompt templates and ensure consistent interpretation by the model. This manual component can hinder scalability and introduce subjective biases into the learning process.}

\m{Looking forward, future work should focus on automating and optimizing the prompt design process. One promising direction is the development of adaptive prompting methods capable of dynamically selecting the most relevant information to include in each iteration based on task performance, uncertainty estimates, or changes in environmental conditions. Automated prompt engineering frameworks, such as AutoPrompt~\cite{autoprompt} and related meta-learning approaches, could be leveraged to discover optimal prompting strategies without human intervention. By coupling such methods with mechanisms for intelligent data selection—prioritizing the most informative sensory-motor samples and performance summaries—the overall efficiency and autonomy of LLM-driven control systems could be further enhanced. The framework would also benefit from the use of more advanced language models and could be extended to dual-agent architectures, in which one model generates control strategies while a second model critiques them, identifies weaknesses, and provides constructive feedback~\cite{ruleguided,selfreflection}.}

\m{Another promising research direction involves identifying the knowledge and skills that are specifically relevant to embodied and situated control, as well as assessing the extent to which existing LLMs possess these capabilities. Plausible candidates include the ability to handle incomplete problem descriptions and to model the physical world.}

\m{Ultimately, overcoming these limitations will be key to transforming LLM-based control from a proof of concept into a scalable and generalizable paradigm for embodied intelligence.}

\bibliographystyle{unsrt}  
\bibliography{main}

\section*{Acknowledgements}

This work was carried out using computational resources provided by FAPESC – Foundation for Research and Innovation Support of the State of Santa Catarina, under grant number 2024TR000090. JTC acknowledges support from the Brazilian National Research Council (CNPq) through grant No. 200013/2025-4. JTC and SN acknowledge support from the European Research Program “Future Artificial Intelligence Research – FAIR” (CUP B53C22003630006, Project Code MIUR PE00000013).

\section*{Author contributions statement}

JTC AND SN conceived the experiment(s),  JTC conducted the experiment(s), and JTC and SN analyzed the results.  Both authors prepared and reviewed the manuscript. 

\section*{Additional information}

Code for replicating the experiments and supplementary material are available at https://github.com/jtyska/llm-robotics-article/; The authors declare no competing interests.

\appendix

\section*{Appendix}

\section*{A. Prompts}
\label{app:prompt-details}
\begin{figure}[!htbp]
    \centering
    \includegraphics[width=\linewidth]{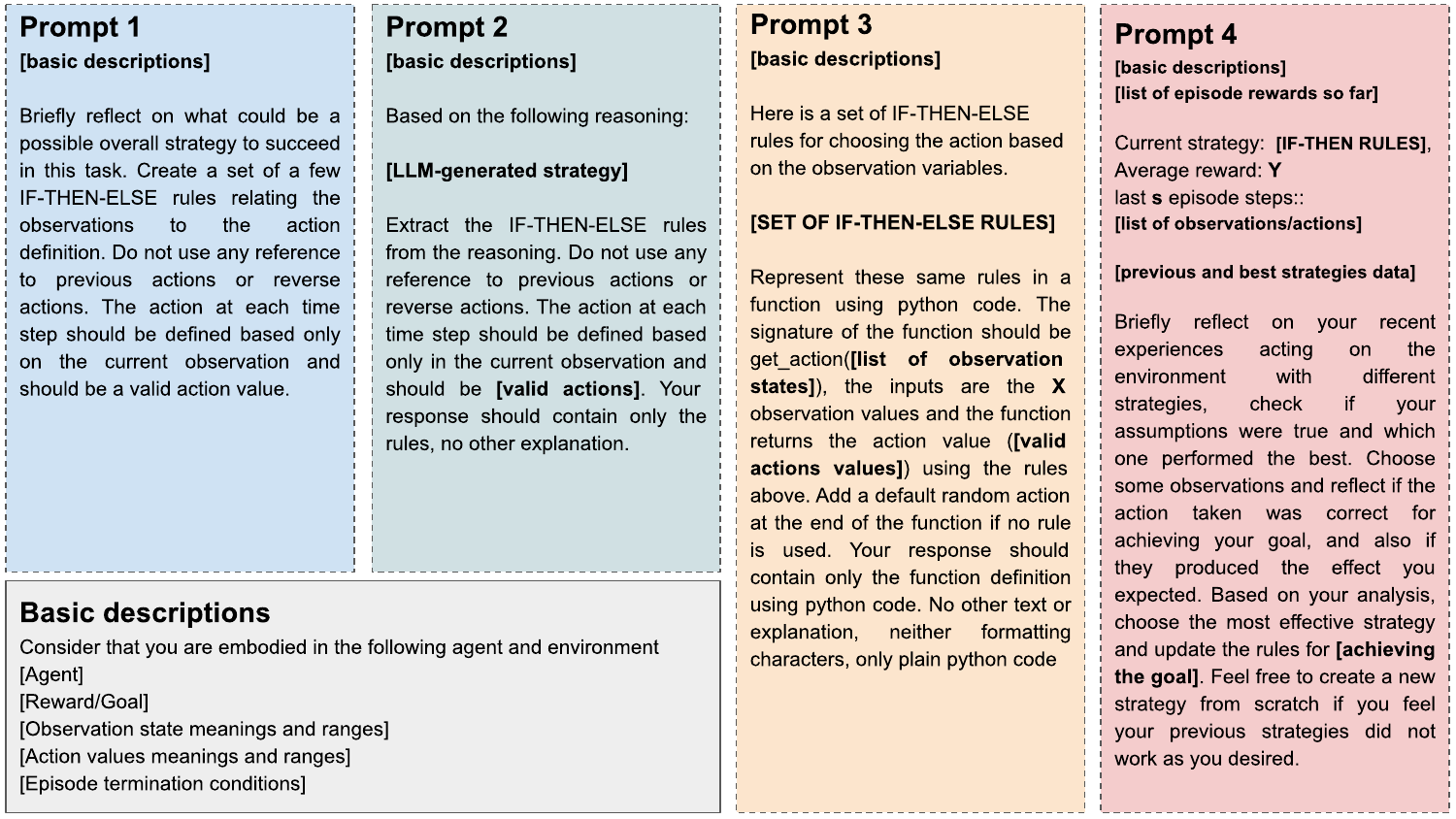} 
    \caption{Prompt used in the experiments. The bracketed text indicates problem-specific text, while the bold text denotes model-generated response to earlier prompts.}
    \label{fig:prompt-details}
\end{figure}

Figure~\ref{fig:prompt-details} presents the four prompts used to query the LLMs initially and during the iterated learning process. The content used to populate the prompt variables—namely [Agent], [Goal and Reward], [Observation Vector], [Action Vector], and [Termination Conditions]—is included in the experimental dataset, which is publicly available at: https://jtyska.github.io/llm-robotics-article/. This repository provides access to the complete dataset generated across all experimental conditions and all replications.

When the Python controller code generated by the LLMs results in execution errors, the models are informed of the issues and given up to ten opportunities to revise the code accordingly.

\section*{B. Tasks}
\label{app:environments-details}

\begin{figure}[!htbp]
\centering
\includegraphics[width=\linewidth]{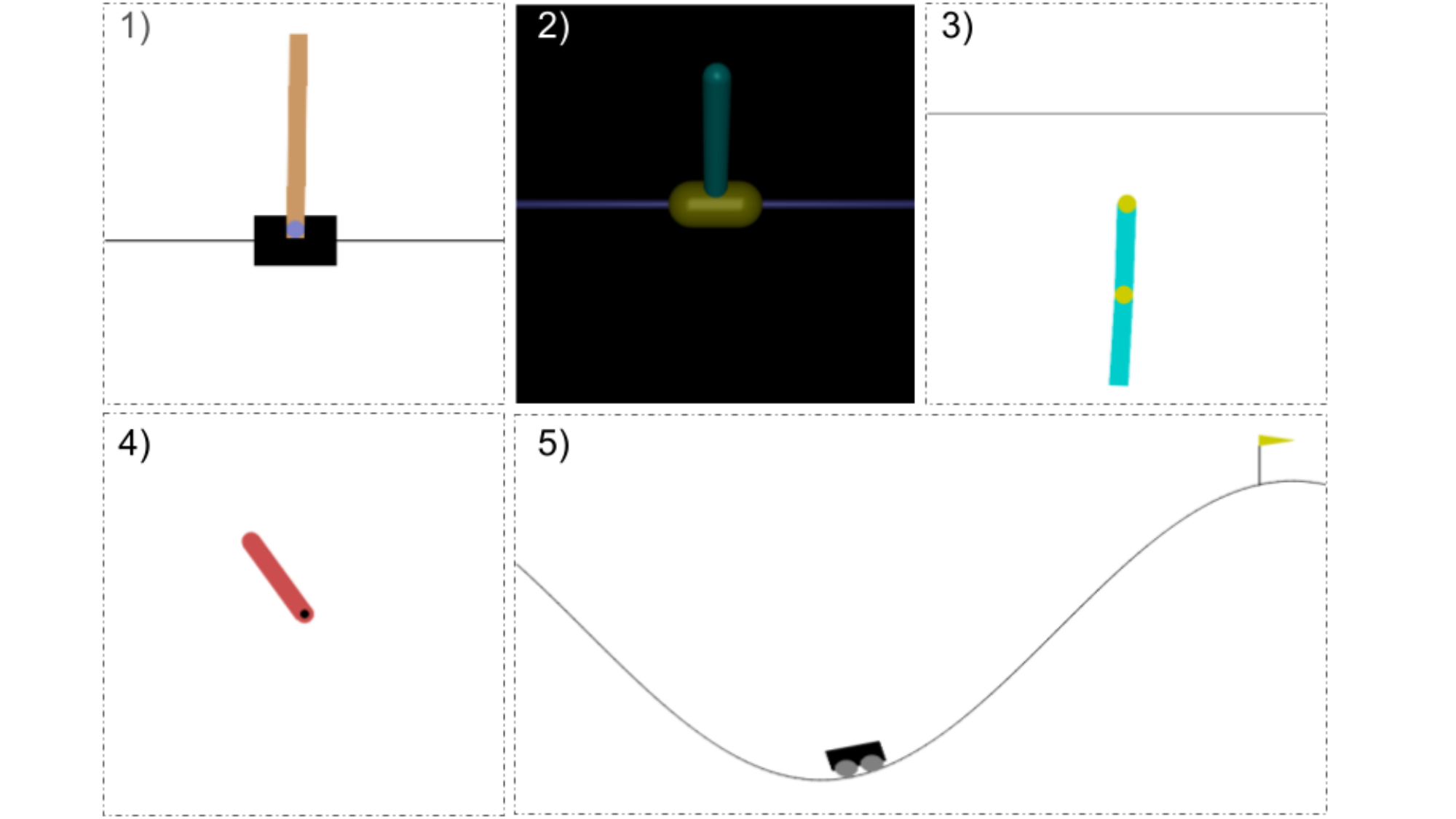}
\caption{Gymnasium problems used in our experiments. a) cart pole, b) inverted pendulum, c) acrobot, d) pendulum, e) mountain car (discrete and continuous version)}
\label{fig:tasks}
\end{figure}

This section describes the tasks considered in the study. The corresponding agent morphologies are illustrated in Figure~\ref{fig:tasks}.

\vspace{0.5em}
\noindent \textbf{1. Cart Pole}

The CartPole task (CartPole-V1 in Gymnasium) involves balancing a pole attached to an unactuated joint atop a moving cart on a frictionless track. The agent can apply discrete forces-left or right-to counteract the pole movements and maintain its upright position for as long as possible, up to a maximum of 500 time steps. 

The observation space is a 4-dimensional vector consisting of: cart position (-4.8 m to 4.8m), cart velocity (m/s), pole angle  (-0.418 rad to 0.418 rad), and pole angular velocity (rad/s). The action space consists of a binary value indicating the force applied to the left (0) or right (1). An episode terminates early if the cart's position exceeds ±2.4 m or the pole angle exceeds ±0.2095 rad.

In our CartPole*1 variant, the action encoding was changed from \{0, 1\} to \{1, 2\} to eliminate the possibility of solving the task using memorized solutions. The CartPole*2 variant further modifies the task by normalizing all observation values to the range [-50 to 50]. 

\vspace{0.5em}
\noindent \textbf{2. Inverted pendulum}

This task, similar in morphology to CartPole, involves stabilizing an unactuated pole on a cart using continuous force control. The agent can apply continuous forces in the range [-3.0, 3.0] N and is rewarded +1 for every time step the pole remains within a specified angular range. The objective is to bring the pole upright as fast as possible and to keep it there as long as possible. 

The observation state is a 4-dimensional vector encoding: cart position (m), pole angle (rad), cart velocity (m/s), and pole angular velocity (rad/s). The action space is a continuous value in the range \([-3.0, 3.0]\) N. Evaluation episodes can last up to 1,000 time steps and terminate early if the pole angle exceeds ±0.2 rad.

\vspace{0.5em}
\noindent \textbf{3. Acrobot}

The Acrobot consists of two connected bars: the first connected to a fixed pivot via an actuated joint, and the second connected to the end of the first via an unactuated joint. The agent applies fixed-intensity torque in the actuated joint, with the objective of raising the tip of the second bar above a specified height. 

At each time step in which the goal has not been achieved, the agent receives a reward of \(-1\). Episodes terminate either when the target height is reached or when the cumulative reward falls below -100.

The observation vector contains six values: the cosine and sine of the angles \(\theta_1\) and \(\theta_2\) normalized in the range \([-1, 1]\)) as well as the angular velocities of the two bars. Specifically, the angular velocities range are: (\(-12.567 \leq \dot{\theta}_1 \leq 12.567\), \(-28.274 \leq \dot{\theta}_2 \leq 28.274\)). \(\theta_1\). Here \(\theta_1\) represents the angle of the first bar relative to the downward vertical, and \(\theta_2\) the angle of the second bar relative to the first.

\vspace{0.5em}
\noindent \textbf{4. Pendulum}

The Pendulum task involves swinging and balancing a bar connected to a fixed pivot via an actuated joint. The agent must bring the bar to an upright position and stabilize it. The reward function is:\(-(\theta^2 + 0.1 \cdot \dot{\theta}^2 + 0.001 \cdot \text{torque}^2)\), where \(\theta\) represents the angle of the bar normalized to the range \([- \pi, \pi]\) (with \(0\) indicating the upright position).

Evaluation episodes last for up to 200 time steps. The observation state is a 3-dimensional vector encoding: x and y coordinates of the free end of the bar—given by (\(x = \cos(\theta)\), \(y = \sin(\theta)\))— and the bar angular velocity (\(\dot{\theta}\)). The ranges of these variables are \(x \in [-1.0, 1.0]\), \(y \in [-1.0, 1.0]\), and \(\dot{\theta} \in [-8.0, 8.0]\). 

The action space comprises a single continuous variable representing the torque applied by the actuator, which can range from -2.0 to 2.0. 

\vspace{0.5em}
\noindent \textbf{5. Mountain car}

The Mountain Car task requires an agent to drive a car from the bottom of a valley to the top of the right hill by applying forces either left or right. The agent receives a reward of -1 for each time step until the car reaches a position $\geq$0.5 meters.

The observation vector includes: car’s position (m), and car's velocity (m/s). 

In the discrete version of the task, the agent chooses among three actions: apply a force left (0), do nothing (1), or apply a force right (2). Episodes last up to 200 steps. 

In the continuous version, the agent selects a continuous force value in [-1.0,1.0]. Episodes in this version may run for up to 1,000 time steps.

\vspace{0.5em}

\section*{C. Selected Interactions}

The following excerpt presents portions of the prompts and responses generated by the Qwen2.5 language model during the 8th replication of the CartPole *2 task, conducted at the optimal temperature setting. The complete dataset is publicly accessible via the webpage: https://jtyska.github.io/llm-robotics-article/. The selected excerpts include the initial iteration and several key stages of the iterative learning process that ultimately resulted in the development of an optimal control strategy. Text in bold denotes the model’s responses to preceding prompts. A subsection title has been added for clarity. Underlined text marks missing sections.

\subsection*{Iteration 0: Prompt 1}

The task description is: Consider that you are embodied in the following agent and environment. 
\\
\\
Agent: A cart over a track that has an unactuated pole attached to it. A force can be applied to the cart, left or right, to counteract the passive movement of the pole.
\\
\\
Goal: Keep the pole balanced upright as long as possible to a maximum of 500 time steps.
\\
\\
Observation state: A 4D vector in the following order:
\\
\\
Cart position, Cart velocity, Pole angle, Pole angular velocity
\\
\\
All observation variables are integer values between -50 and 50;
\\
\\
Action: Discrete:\\
1=move left(1)\\
2=move right(2)\\        
\\
Negative values mean left direction, positive values right direction.
\\
\\
Failure Conditions:\\ 
- Pole angle exceeds the range [-25,25]\\
- Cart position exceeds the range [-25,25]
\\
\\
Briefly reflect on what could be a possible overall strategy to succeed in this task. Create a set of a few IF-THEN-ELSE rules relating the observations to the action definition. Do not use any reference to previous actions or reverse actions. The action at each time step should be defined based only on the current observation and should be either left(1) or right(2).
\\
\\
\begin{BoldVerbatim}
To succeed in this task of keeping the pole balanced upright for as long as possible, a strategy 
can be designed using simple IF-THEN-ELSE rules that respond to the current state of the cart and 
the pole. The goal is to apply forces that counteract the movement that would cause the pole to fall 
or the cart to move out of bounds.

Here's a set of IF-THEN-ELSE rules for this task:

1. **If the Pole Angle is Positive (leaning right):**
   - **Then** Move Right (2)
   
2. **Else If the Pole Angle is Negative (leaning left):**
   - **Then** Move Left (1)

3. **If the Cart Position is Near the Right Boundary:**
   - **Then** Move Left (1)
   
4. **Else If the Cart Position is Near the Left Boundary:**
   - **Then** Move Right (2)

5. **If the Pole Angular Velocity is Positive and High:**
   - **Then** Move Right (2) to counteract rapid rightward movement
   
6. **Else If the Pole Angular Velocity is Negative and High:**
   - **Then** Move Left (1) to counteract rapid leftward movement

7. **If the Cart Velocity is Positive and High:**
   - **Then** Move Left (1) to slow down the cart's rightward motion
   
8. **Else If the Cart Velocity is Negative and High:**
   - **Then** Move Right (2) to slow down the cart's leftward motion

9. **If the Pole Angle is within a Small Range of Zero (e.g., -5 to 5):**
   - **Then** Apply a small corrective force based on the Cart Velocity:
     - If Cart Velocity is Positive, Move Left (1)
     - If Cart Velocity is Negative, Move Right (2)

These rules aim to keep the pole balanced by responding directly to its angle and angular velocity
while also managing the cart's position and velocity to avoid boundary conditions. The strategy
balances immediate reaction to the pole's state with prevention of excessive movement that could 
lead to failure.

\end{BoldVerbatim}

\subsection*{Iteration 0: Prompt 2}

The task description is: Consider that you are embodied in the following agent and environment. 
        
Agent: A cart over \textbf{\underline{[… CONTINUE AS ABOVE]}}
\\
\\
Based on the following reasoning
\\
\\
To succeed in this task of keeping the pole balanced upright for as long as possible,
a strategy can be designed using simple IF-THEN-ELSE rules that respond to the current state of the cart and the pole. The goal is to apply forces that counteract the movement that would cause the pole to fall or the cart to move out of bounds.
\\
\\
Here's a set of IF-THEN-ELSE rules for this task:
\\
1. **If the Pole Angle is Positive (leaning right):**\\
\hspace*{2em}- **Then** Move Right (2)
\\
2. **Else If the Pole Angle is Negative (leaning left):**\\
\hspace*{2em}- **Then** Move Left (1)
\\
3. **If the Cart Position is Near the Right Boundary:**\\
\hspace*{2em}- **Then** Move Left (1)
\\
4. **Else If the Cart Position is Near the Left Boundary:**\\
\hspace*{2em}- **Then** Move Right (2)
\\
5. **If the Pole Angular Velocity is Positive and High:**\\
\hspace*{2em}- **Then** Move Right (2) to counteract rapid rightward movement
\\
6. **Else If the Pole Angular Velocity is Negative and High:**\\
\hspace*{2em}- **Then** Move Left (1) to counteract rapid leftward movement
\\
7. **If the Cart Velocity is Positive and High:**\\
\hspace*{2em}- **Then** Move Left (1) to slow down the cart's rightward motion
\\
8. **Else If the Cart Velocity is Negative and High:**\\
\hspace*{2em}- **Then** Move Right (2) to slow down the cart's leftward motion
\\
9. **If the Pole Angle is within a Small Range of Zero (e.g., -5 to 5):**\\
\hspace*{2em}- **Then** Apply a small corrective force based on the Cart Velocity:\\
\hspace*{3em}- If Cart Velocity is Positive, Move Left (1)\\
\hspace*{3em}- If Cart Velocity is Negative, Move Right (2)
\\
\\
These rules aim to keep the pole balanced by responding directly to its angle and angular velocity while also managing the cart's position and velocity to avoid boundary conditions. The strategy balances immediate reaction to the pole's state with prevention of excessive movement that could lead to failure.
\\
\\
\vspace*{1em}
Extract the IF-THEN-ELSE rules from the reasoning. Do not use any reference to previous actions or reverse actions. The action at each time step should be defined based only in the current observation and should be either left(1) or right(2). Your response should contain only the rules, no other explanation.
\\
\begin{BoldVerbatim}
1. If Pole Angle > 0 Then Move Right (2)
2. Else If Pole Angle < 0 Then Move Left (1)
3. If Cart Position >= 20 Then Move Left (1)
4. Else If Cart Position <= -20 Then Move Right (2)
5. If Pole Angular Velocity > 10 Then Move Right (2)
6. Else If Pole Angular Velocity < -10 Then Move Left (1)
7. If Cart Velocity > 10 Then Move Left (1)
8. Else If Cart Velocity < -10 Then Move Right (2)
9. If Pole Angle >= -5 And Pole Angle <= 5
   - If Cart Velocity > 0 Then Move Left (1)
   - If Cart Velocity < 0 Then Move Right (2)

\end{BoldVerbatim}

\subsection*{Iteration 0: Prompt 3}

The task description is \textbf{\underline{[…. CONTINUE AS ABOVE]}}\\

Here are a set of IF-THEN-ELSE rules for choosing the action based on the observation variables.
\\
\\
1. If Pole Angle > 0 Then Move Right (2)\\
2. Else If Pole Angle < 0 Then Move Left (1)\\
3. If Cart Position >= 20 Then Move Left (1)\\
4. Else If Cart Position <= -20 Then Move Right (2)\\
5. If Pole Angular Velocity > 10 Then Move Right (2)\\
6. Else If Pole Angular Velocity < -10 Then Move Left (1)\\
7. If Cart Velocity > 10 Then Move Left (1)\\
8. Else If Cart Velocity < -10 Then Move Right (2)\\
9. If Pole Angle >= -5 And Pole Angle <= 5\\
\hspace*{2em}- If Cart Velocity > 0 Then Move Left (1)\\
\hspace*{2em}- If Cart Velocity < 0 Then Move Right (2)
\\
\\
Represent these same rules in a function using python code. The signature of the function should be get\_action(cart\_position, cart\_velocity, pole\_angle,pole\_angular\_velocity), the inputs are the four observation values and the function returns the action value (1 for left; or 2 for right) using the rules above. Add a default random action (return random.randint(1, 2)) at the end of the function if no rule is used. Your response should contain only the function definition using python code. No other text or explanation, neigher formatting characters, only plain python code.
\\  
\\

\begin{lstlisting}[language=Python]
import random

def get_action(cart_position, cart_velocity, pole_angle, pole_angular_velocity):
    if pole_angle > 0:
        return 2
    elif pole_angle < 0:
        return 1
    if cart_position >= 20:
        return 1
    elif cart_position <= -20:
        return 2
    if pole_angular_velocity > 10:
        return 2
    elif pole_angular_velocity < -10:
        return 1
    if cart_velocity > 10:
        return 1
    elif cart_velocity < -10:
        return 2
    if -5 <= pole_angle <= 5:
        if cart_velocity > 0:
            return 1
        elif cart_velocity < 0:
            return 2
    return random.randint(1, 2)
\end{lstlisting}

\hspace*{-1.5em}Trial 1: Reward = 49.0\\
Trial 1: Reward = 47.0\\
Trial 1: Reward = 46.0\\
Trial 1: Reward = 39.0\\
Trial 1: Reward = 57.0\\
Trial 1: Reward = 58.0\\
Trial 1: Reward = 35.0\\
Trial 1: Reward = 47.0\\
Trial 1: Reward = 38.0\\
Trial 1: Reward = 56.0\\
Trial 1: Reward = 34.0\\
Trial 1: Reward = 40.0\\
Trial 1: Reward = 51.0\\
Trial 1: Reward = 46.0\\
Trial 1: Reward = 44.0\\
Trial 1: Reward = 77.0\\
Trial 1: Reward = 55.0\\
Trial 1: Reward = 65.0\\
Trial 1: Reward = 64.0\\
Trial 1: Reward = 49.0\\

\subsection*{ITERATION 0: Prompt 4}

The task description is: \textbf{\underline{[….CONTUNUE AS ABOVE]}}

Your experience with this strategy was: \textbf{\underline{[….CONTUNUE AS ABOVE]}}

Briefly reflect on your recent experiences acting on the environment with different strategies, check if your assumptions were true and which one performed the best. Choose some observations and reflect if the action taken was correct for achieving your goal, and also if they produced the effect you expected. Based on your analysis, choose the most effective strategy and update the rules for keeping the pole balanced for longer. Feel free to create a new strategy from scratch if you feel your previous strategies did not work as you desired.
\\
\\
\begin{BoldVerbatim}
The current strategy aims to balance the pole by adjusting the cart's position and velocity
based on the pole angle, angular velocity, and cart position. However, it seems that this
strategy is not consistently effective in keeping the pole balanced for a longer duration. 
Let's break down some observations from the last 20 steps of a trial:

1. **Observation: [0 16 -7 -23]; Action: Move Left (1)**
   - The cart was at position 0 with a velocity of 16, and the pole angle was -7 with an angular 
   velocity of -23.
   - The action to move left is appropriate because the pole is tilting to the right, but the high
   negative angular velocity suggests that the pole is moving quickly to the left. This might lead
   to instability. 
2. **Observation: [0 10 -12 -14]; Action: Move Left (1)**
   - The cart was at position 0 with a velocity of 10, and the pole angle was -12 with an angular
   velocity of -14.
   - Moving left is correct as the pole is tilting to the right, but the high negative angular
   velocity indicates that the pole is moving quickly to the left. The cart might not be able to
   counteract this movement effectively.

3. **Observation: [0 -22 -5 29]; Action: Move Left (1)**
   - The cart was at position 0 with a velocity of -22, and the pole angle was -5 with an angular
   velocity of 29.
   - Moving left is incorrect because the pole is tilting to the left, and the cart should move right
   to counteract this. Additionally, the high positive angular velocity suggests that the pole is
   moving quickly to the right. 

4. **Observation: [-2 -13 25 17]; Action: None**
   - The cart was at position -2 with a velocity of -13, and the pole angle was 25 with an angular
   velocity of 17.
   - No action is taken here, which is problematic because the pole angle is already at the failure
   condition. An immediate corrective action should be taken to move right.

### Updated Strategy

Based on the analysis, the current strategy has some weaknesses in handling high angular velocities
and edge cases where the cart or pole are close to their limits. Here's an updated strategy that
addresses these issues:

1. **Pole Angle and Angular Velocity:**
   - If Pole Angle > 0 AND Pole Angular Velocity > 5 Then Move Right (2)
   - Else If Pole Angle < 0 AND Pole Angular Velocity < -5 Then Move Left (1)
   - If Pole Angle > 0 AND Pole Angular Velocity <= 5 Then Move Left (1)
   - Else If Pole Angle < 0 AND Pole Angular Velocity >= -5 Then Move Right (2)

2. **Cart Position:**
   - If Cart Position >= 20 THEN Move Left (1)
   - Else If Cart Position <= -20 THEN Move Right (2)

3. **Pole Near Failure Conditions:**
   - If Pole Angle > 20 OR Pole Angle < -20 THEN Move in the direction opposite to the pole angle 
   (to correct quickly)
     - If Pole Angle > 20 Then Move Left (1)
     - Else If Pole Angle < -20 Then Move Right (2)

4. **Cart Velocity:**
   - If Cart Velocity > 15 Then Move Left (1)
   - Else If Cart Velocity < -15 Then Move Right (2)

5. **Fine-Tuning for Small Angles:**
   - If Pole Angle >= -5 AND Pole Angle <= 5
     - If Cart Velocity > 0 Then Move Left (1)
     - If Cart Velocity < 0 Then Move Right (2)
   - If Pole Angular Velocity > 10 THEN Move in the direction opposite to the pole angle (to correct 
   quickly)
     - If Pole Angular Velocity > 10 Then Move Left (1)
     - Else If Pole Angular Velocity < -10 Then Move Right (2)

### Rationale
- **Pole Angle and Angular Velocity:** This rule helps in stabilizing the pole when it is tilting
significantly or moving rapidly.
- **Cart Position:** Ensures the cart does not go out of bounds.
- **Pole Near Failure Conditions:** Immediate corrective actions to prevent failure.
- **Cart Velocity:** Helps in managing the cart's momentum to avoid overshooting.
- **Fine-Tuning for Small Angles:** Smaller adjustments are made when the pole is nearly balanced
to maintain stability.

This updated strategy should help in keeping the pole balanced for a longer duration and improve the
average reward.
\end{BoldVerbatim}

\subsection*{Iteration 2: Prompt 3}
\vspace*{1em}
The task description is: \textbf{\underline{[….CONTINUE AS ABOVE]}}
\\
\\
Your current overall strategy was \textbf{\underline{[….CONTINUE AS ABOVE]}}
\\
\\
Last 20 steps from a trial using this strategy. The average reward in 20 trials was 49.85/500 
    
[  0  16  -7 -23];1 \textbf{\underline{[….CONTINUE AS ABOVE]}}
\\
\\                  
Briefly reflect on your recent experiences acting on the environment with different strategies, check if your assumptions were true and which one performed the best. Choose some observations and reflect if the action taken was correct for achieving your goal, and also if they produced the effect you expected. Based on your analysis, choose the most effective strategy and update the rules for keeping the pole balanced for longer. Feel free to create a new strategy from scratch if you feel your previous strategies did not work as you desired.
\\
\\
Your previous reasoning generated an invalid action, it should be 1 (left) or 2 (right) but it was None
\begin{BoldVerbatim}
### Reflection on the Recent Experiences

#### **Analysis of the Strategy**

The strategy I used attempted to balance the pole by considering multiple factors such as pole 
angle, angular velocity, cart position, and cart velocity. However, the performance of this 
strategy was suboptimal, with an average reward of only 49.85 out of 500 time steps in 20 trials. 
This indicates that the strategy often failed to keep the pole balanced for a significant duration.

#### **Key Observations and Actions**

Let's analyze some key observations from the last 20 steps:

1. **Step 7:**
   - **Observation:** [ 1, -3, -17, 3]
   - **Action Taken:** Move Left (1)
   - **Expected Effect:** The pole is tilting to the left with a negative angle (-17) and a small 
   positive angular velocity (3). Moving left should help counteract this tilt.
   - **Actual Effect:** The cart moved further left, but the pole's angle continued to increase in 
   magnitude.

2. **Step 14:**
   - **Observation:** [0, -26, -1, 33]
   - **Action Taken:** Move Left (1)
   - **Expected Effect:** The cart is moving very fast to the left (-26) with a large positive angular 
   velocity (33. Moving left should slow down the cart and help stabilize the pole.
   - **Actual Effect:** The cart continued to move left, but the pole's angle became more positive.

3. **Step 19:**
   - **Observation:** [-2, -16, 22, 21]
   - **Action Taken:** Move Left (1)
   - **Expected Effect:** The pole is tilting to the right with a positive angle (22) and a positive 
   angular velocity (21). Moving left should help counteract this tilt.
   - **Actual Effect:** The cart moved further left, but the pole's angle increased in magnitude.

4. **Step 20:**
   - **Observation:** [-2, -13, 25, 17]
   - **Action Taken:** None
   - **Expected Effect:** This is an invalid action as it should be either 1 (left) or 2 (right).
   - **Actual Effect:** The simulation likely terminated due to this invalid action.

### **Reflection on the Strategy**

The strategy was overly complex and sometimes contradictory. For example, when the pole angle and 
angular velocity were both negative, the strategy suggested moving left, which might not always be the 
best course of action. Additionally, the conditions for cart position and velocity were too strict and
often led to unnecessary movements that destabilized the system.

### **Updated Strategy**

To improve the performance, I will simplify the strategy by focusing on the most critical factors: 
pole angle and angular velocity. The goal is to counteract the pole's movement while maintaining 
a stable cart position.

#### **New Strategy:**

1. **Pole Angle and Angular Velocity:**
   - If **Pole Angle > 5** or **Pole Angular Velocity > 10**, Move Right (2).
   - If **Pole Angle < -5** or **Pole Angular Velocity < -10**, Move Left (1).

2. **Cart Position:**
   - If **Cart Position >= 20**, Move Left (1) to avoid the cart going out of bounds.
   - If **Cart Position <= -20**, Move Right (2) to avoid the cart going out of bounds.

3. **Default Action:**
   - If none of the above conditions are met, maintain a balanced action:
     - If **Pole Angle > 0** and **Pole Angular Velocity < 5**, Move Left (1).
     - If **Pole Angle < 0** and **Pole Angular Velocity > -5**, Move Right (2).

#### **New Strategy Rules:**

1. **If Pole Angle > 5 or Pole Angular Velocity > 10:** Move Right (2)
2. **Else if Pole Angle < -5 or Pole Angular Velocity < -10:** Move Left (1)
3. **Else if Cart Position >= 20:** Move Left (1)
4. **Else if Cart Position <= -20:** Move Right (2)
5. **Else if Pole Angle > 0 and Pole Angular Velocity < 5:** Move Left (1)
6. **Else if Pole Angle < 0 and Pole Angular Velocity > -5:** Move Right (2)

### **Conclusion**

The new strategy is designed to be more straightforward and focused on the most critical factors for 
balancing the pole. By reducing the complexity and ensuring that actions are taken based on the primary
conditions, we should see improved performance in keeping the pole balanced for a longer duration.    
\end{BoldVerbatim}

\subsection*{Iteration 3: Prompt 3}
\vspace*{1em}
The task description is: \textbf{\underline{[….CONTINUE AS ABOVE]}}
\\
\\
You already tried 2 different strategies , and the sequence of rewards was: [49.85, 116.1]
\\
\\
Your current overall strategy was this one:
\begin{verbatim}
1. **If Pole Angle > 5 or Pole Angular Velocity > 10:** Move Right (2)
2. **Else if Pole Angle < -5 or Pole Angular Velocity < -10:** Move Left (1)
3. **Else if Cart Position >= 20:** Move Left (1)
4. **Else if Cart Position <= -20:** Move Right (2)
5. **Else if Pole Angle > 0 and Pole Angular Velocity < 5:** Move Left (1)
6. **Else if Pole Angle < 0 and Pole Angular Velocity > -5:** Move Right (2) 
\end{verbatim}
Your experience with this strategy was:
\begin{verbatim}
Last 20 steps from a trial using this strategy. The average reward in 20 trials was 116.1/500
                
[-1 -3  9  6];2
[-1  0 10  1];1
[-1 -3 10  6];2	
[-1  0 11  2];1
[-1 -3 12  7];2
[-1  0 13  3];1
[-1 -3 13  8];2
[-1  0 14  4];1
[-1 -3 15 10];2
[-1  0 16  5];1
[-1 -3 17 11];2
[-1  0 19  7];2
[-1  3 20  3];1
[-1  0 20  9];2
[-1  3 21  5];1
[-1  0 22 10];2
[-1  3 23  6];2
[-1  6 24  3];1
[-1  3 25  9];2
[-1  6 26  5];None
\end{verbatim}

Your previous overall strategy was this one:

\begin{verbatim}
IF Pole Angle > 0 AND Pole Angular Velocity > 5 THEN Move Right (2)
ELSE IF Pole Angle < 0 AND Pole Angular Velocity < -5 THEN Move Left (1)
ELSE IF Pole Angle > 0 AND Pole Angular Velocity <= 5 THEN Move Left (1)
ELSE IF Pole Angle < 0 AND Pole Angular Velocity >= -5 THEN Move Right (2)

IF Cart Position >= 20 THEN Move Left (1)
ELSE IF Cart Position <= -20 THEN Move Right (2)

IF Pole Angle > 20 THEN Move Left (1)
ELSE IF Pole Angle < -20 THEN Move Right (2)

IF Cart Velocity > 15 THEN Move Left (1)
ELSE IF Cart Velocity < -15 THEN Move Right (2)

IF Pole Angle >= -5 AND Pole Angle <= 5
THEN IF Cart Velocity > 0 THEN Move Left (1)
ELSE IF Cart Velocity < 0 THEN Move Right (2)

IF Pole Angular Velocity > 10 THEN Move Left (1)
ELSE IF Pole Angular Velocity < -10 THEN Move Right (2)
\end{verbatim}

               Your experience with this strategy was:

Last 20 steps from a trial using this strategy. The average reward in 20 trials was 49.85/500  \textbf{\underline{[….CONTINUE AS ABOVE]}}
\\
\\
Briefly reflect on your recent experiences acting on the environment with different strategies, check if your assumptions were true and which one performed the best. Choose some observations and reflect if the action taken was correct for achieving your goal, and also if they produced the effect you expected. Based on your analysis, choose the most effective strategy and update the rules for keeping the pole balanced for longer. Feel free to create a new strategy from scratch if you feel your previous strategies did not work as you desired.

\begin{BoldVerbatim}
                
**Observations from Last 20 Steps:**
- The cart position was consistently at -1, indicating it did not move much.
- The pole angle increased steadily from 9 to 26, suggesting the strategy failed to keep the pole
balanced.
- The actions taken were mostly moving right (2), which did not effectively counteract the increasing 
pole angle.

**Analysis:**
- The strategy is overly dependent on the pole angle and angular velocity thresholds. When these 
conditions are met, it applies a strong corrective action (move left or right).
- However, when the pole angle and angular velocity are within certain ranges, the actions taken were 
not sufficient to correct the balance.
- The cart position constraints (3 and 4) did not come into play much because the cart was already 
near the boundary. 
#### Previous Strategy Analysis:
1. **If Pole Angle > 0 AND Pole Angular Velocity > 5:** Move Right (2)
2. **Else if Pole Angle < 0 AND Pole Angular Velocity < -5:** Move Left (1)
3. **Else if Pole Angle > 0 AND Pole Angular Velocity <= 5:** Move Left (1)
4. **Else if Pole Angle < 0 AND Pole Angular Velocity >= -5:** Move Right (2)
5. **If Cart Position >= 20:** Move Left (1)
6. **Else if Cart Position <= -20:** Move Right (2)
7. **If Pole Angle > 20:** Move Left (1)
8. **Else if Pole Angle < -20:** Move Right (2)
9. **If Cart Velocity > 15:** Move Left (1)
10. **Else if Cart Velocity < -15:** Move Right (2)
11. **If Pole Angle >= -5 AND Pole Angle <= 5:**
    - If Cart Velocity > 0: Move Left (1)
    - Else if Cart Velocity < 0: Move Right (2)
12. **If Pole Angular Velocity > 10:** Move Left (1)
13. **Else if Pole Angular Velocity < -10:** Move Right (2)

**Observations from Last 20 Steps:**
- The cart position decreased steadily, and the pole angle became increasingly negative.
- The actions taken were predominantly moving left (1), which did not effectively balance the pole.

**Analysis:**
- This strategy is more complex and includes multiple conditions to ensure the cart and pole are
balanced.
- However, it may be too reactive and not proactive enough. The frequent left movements might have
exacerbated the imbalance rather than corrected it.

### Updated Strategy

Based on the reflections, I propose a new strategy that balances reactivity with proactivity:

1. **If Pole Angle > 5 or Pole Angular Velocity > 10:** Move Right (2)
2. **Else if Pole Angle < -5 or Pole Angular Velocity < -10:** Move Left (1)
3. **Else if Cart Position >= 18:** Move Left (1)
4. **Else if Cart Position <= -18:** Move Right (2)
5. **If Pole Angle > 3 and Pole Angular Velocity > 0:** Move Left (1) with a slight delay 
to avoid overreacting
6. **Else if Pole Angle < -3 and Pole Angular Velocity < 0:** Move Right (2) with a slight delay 
to avoid overreacting
7. **If Cart Velocity > 5:** Move Left (1) but only if Pole Angle is not significantly negative
8. **If Cart Velocity < -5:** Move Right (2) but only if Pole Angle is not significantly positive
9. **If Pole Angle is between -3 and 3:**
    - If Pole Angular Velocity > 0: Move Left (1)
    - Else if Pole Angular Velocity < 0: Move Right (2)

### Explanation of the New Strategy:
- The first four rules handle extreme cases for pole angle, angular velocity, and cart position.
- Rules 5 and 6 introduce a slight delay in corrective actions to prevent overreacting when the pole 
is slightly off balance.
- Rules 7 and 8 consider the cart's velocity but with a condition to avoid conflicting movements that
could destabilize the pole.
- Rule 9 ensures that small adjustments are made to maintain balance when the pole is nearly centered.

This strategy aims to be more balanced and adaptive, reducing the risk of overcorrection while ensuring
the pole remains stable.

\end{BoldVerbatim}

\subsection*{Iteration 4: Prompt 3}
\vspace*{1em}
The task description is: \textbf{\underline{[….CONTINUE AS ABOVE]}}

\textbf{\underline{[MISSING TEXT]}}
\\
\\
Extract the IF-THEN-ELSE rules from the reasoning. Do not use any reference to previous actions or reverse actions. The action at each time step should be defined based only in the current observation and should be either left(1) or right(2). Your response should contain only the rules, no other explanation.
\\
\\
\begin{BoldVerbatim}
IF Pole Angle > 3 OR Pole Angular Velocity > 5 THEN Move Right (2)
ELSE IF Pole Angle < -3 OR Pole Angular Velocity < -5 THEN Move Left (1)
ELSE IF Cart Position >= 20 AND Pole Angle <= 0 THEN Move Left (1)
ELSE IF Cart Position >= 20 AND Pole Angle > 0 THEN Move Right (2)
ELSE IF Cart Position <= -20 AND Pole Angle >= 0 THEN Move Right (2)
ELSE IF Cart Position <= -20 AND Pole Angle < 0 THEN Move Left (1)
ELSE IF Cart Velocity > 5 OR Cart Velocity < -5 THEN
    IF Pole Angle > 0 THEN Move Right (2)
    ELSE IF Pole Angle < 0 THEN Move Left (1)
ELSE IF Pole Angle > 0 THEN Move Right (2)
ELSE IF Pole Angle < 0 THEN Move Left (1)
\end{BoldVerbatim}
\textbf{\underline{[MISSING TEXT]}}
\\
\\
Trial 4: Reward = 500.0\\
Trial 4: Reward = 500.0\\
Trial 4: Reward = 500.0\\
Trial 4: Reward = 500.0\\
Trial 4: Reward = 500.0\\
Trial 4: Reward = 500.0\\
Trial 4: Reward = 500.0\\
Trial 4: Reward = 500.0\\
Trial 4: Reward = 500.0\\
Trial 4: Reward = 500.0\\
Trial 4: Reward = 500.0\\
Trial 4: Reward = 500.0\\
Trial 4: Reward = 500.0\\
Trial 4: Reward = 500.0\\
Trial 4: Reward = 500.0\\
Trial 4: Reward = 500.0\\
Trial 4: Reward = 500.0\\
Trial 4: Reward = 500.0\\
Trial 4: Reward = 500.0\\
Trial 4: Reward = 500.0\\

\end{document}